%% file: main.tex
\documentclass[runningheads]{llncs}
\usepackage{graphicx}

\usepackage{tikz}
\usepackage{comment}
\usepackage{amsmath,amssymb} 
\usepackage{color}

\usepackage[accsupp]{axessibility}  

\usepackage{graphicx}
\usepackage{amsmath}
\usepackage{amssymb}
\usepackage{booktabs}
\usepackage{epsfig}
\usepackage{overpic}
\usepackage{wrapfig}
\usepackage{mwe}
\usepackage{color}
\usepackage{xcolor}
\usepackage{caption}

\usepackage{subcaption}

\usepackage{adjustbox}
\usepackage{breakcites}

\begin{document}
\pagestyle{headings}
\mainmatter
\def\ECCVSubNumber{6203}  

\title{PatchRD: Detail-Preserving Shape Completion  \\
by Learning Patch Retrieval and Deformation}

\titlerunning{PatchRD}
%
\author{Bo Sun\inst{1} \and
Vladimir G. Kim\inst{2} \and
Noam Aigerman\inst{2} \and
Qixing Huang\inst{1}
\and
Siddhartha Chaudhuri\inst{2,3}
}
\authorrunning{Sun et al.}
%
\institute{UT Austin \and Adobe Research \and IIT Bombay}
\maketitle

\input{00_abstract}


\input{01_intro}

\input{02_related}
\input{03_overview}

\input{04_1Retrieval}

\input{04_2Deformation}

\input{05_results}

\input{06_conclusions}

\clearpage
%
%
\bibliographystyle{plain}
\bibliography{egbib}
\end{document}


\pagestyle{headings}
\mainmatter
\def\ECCVSubNumber{6203}  

\title{Supplementary Materials for \\
PatchRD: Detail-Preserving Shape Completion  \\
by Learning Patch Retrieval and Deformation}

\titlerunning{PatchRD}
%
\author{Bo Sun\inst{1} \and
Vladimir G. Kim\inst{2} \and
Noam Aigerman\inst{2} \and
Qixing Huang\inst{1}
\and
Siddhartha Chaudhuri\inst{2,3}
}
\authorrunning{Sun et al.}
%
\institute{UT Austin \and Adobe Research \and IIT Bombay }
\maketitle

\input{07_supp}

%
%
\bibliographystyle{splncs04}
\bibliography{egbib}

%% file: 00_abstract.tex
\begin{abstract}
This paper introduces a data-driven shape completion approach that focuses on completing geometric details of missing regions of 3D shapes. We observe that existing generative methods lack the training data and representation capacity to synthesize plausible, fine-grained details with complex geometry and topology. Our key insight is to copy and deform patches from the partial input to complete missing regions. This enables us to preserve the style of local geometric features, even if it drastically differs from the training data. Our fully automatic approach proceeds in two stages. First, we learn to retrieve candidate patches from the input shape. Second, we select and deform some of the retrieved candidates to seamlessly blend them into the complete shape. This method combines the advantages of the two most common completion methods: similarity-based single-instance completion, and completion by learning a shape space. We leverage repeating patterns by retrieving patches from the partial input, and learn global structural priors by using a neural network to guide the retrieval and deformation steps. Experimental results show our approach considerably outperforms baselines across multiple datasets and shape categories. Code and data are available at \href{https://github.com/GitBoSun/PatchRD}{https://github.com/GitBoSun/PatchRD}.
\end{abstract}

%% file: 01_intro.tex
\section{Introduction}

Completing geometric objects is a fundamental problem in visual computing with a wide range of applications. For example, when scanning complex geometric objects, it is always difficult to scan every point of the underlying object~\cite{Levoy:2000:DMP}. The scanned geometry usually contains various levels of holes and missing geometries, making it critical to develop high-quality geometry completion techniques~\cite{Turk:94,Curless:1996:VRIP,Wheeler:1998:CSM,Davis:2002,Anguelov:2005:SCAPE,Hu:2012:FH,Guo:2018:HF}. Geometry completion is also used in interactive shape modeling~\cite{Chaudhuri:2010:DDS}, as a way to suggest additional content to add to a partial 3D object/scene. Geometry completion is challenging, particularly when the missing regions contain non-trivial geometric content.

\begin{figure}
\centering
\begin{overpic}[width=0.9\textwidth]{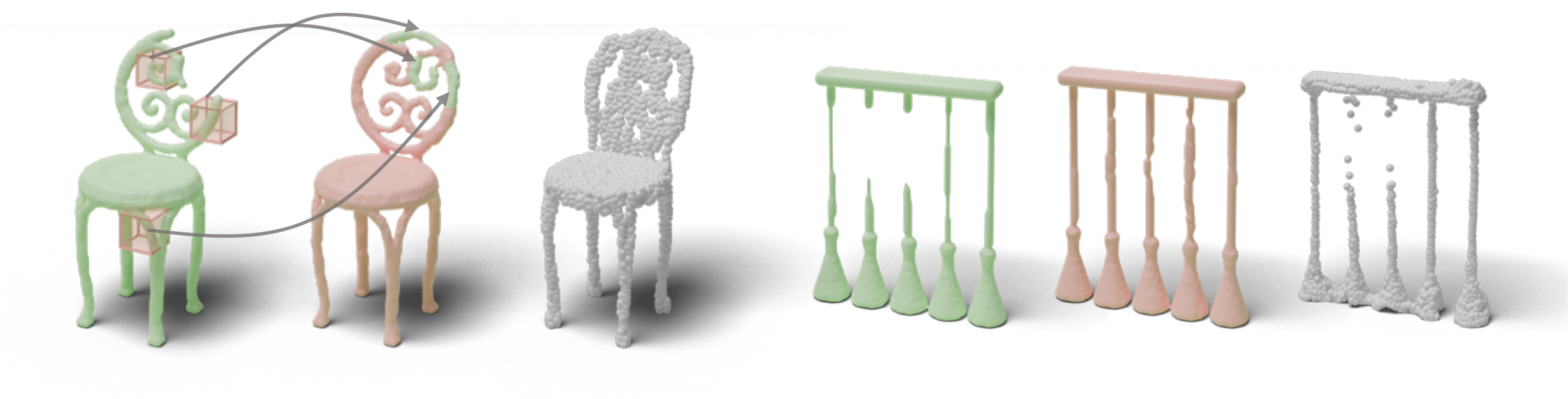}
\put(6,0) {Input }
\put(18,0) {PatchRD }
\put(19,-3) { (Ours) }
\put(32,0) {Generative }
\put(32.5,-3) {Model \cite{xiang21snow}}
\put(54,0) {Input }
\put(67.5,0) {PatchRD }
\put(68.5,-3) { (Ours) }
\put(82,0) {Generative }
\put(82.5,-3) {Model \cite{xiang21snow}}
\end{overpic}
\caption{We propose PatchRD, a non-parametric shape completion method based on patch retrieval and deformation. Compared with the parametric generation methods, our method is able to recover complex geometric details as well as keeping the global shape smoothness. }
\label{Figure:teaser}
\end{figure}
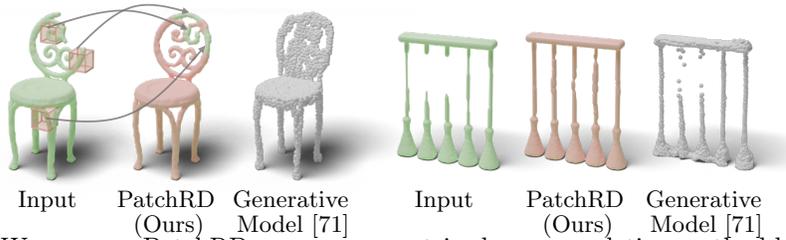

Early geometry completion techniques focus on hole filling~\cite{Turk:94,Curless:1996:VRIP,Wheeler:1998:CSM,Davis:2002,Anguelov:2005:SCAPE,Hu:2012:FH,Guo:2018:HF}. These techniques rely on the assumption that the missing regions are simple surface patches and can be filled by smoothly extending hole regions. Filling regions with complex shapes rely on data priors. Existing approaches fall into two categories. The first category extracts similar regions from the input shape. The hypothesis is that a physical 3D object naturally exhibits repeating content due to symmetries and texture. While early works use user-specified rules to retrieve and fuse similar patches, recent works have studied using a deep network to automatically complete a single image or shape~\cite{DBLP:conf/cvpr/UlyanovVL18,Hanocka:2019:MeshCNN,Hanocka:2020:Point2Mesh}. The goal of these approaches is to use different layers of the neural network (e.g., a convolutional neural network) to automatically extract repeating patterns. However, these approaches are most suitable when the repeating patterns are prevalent within the partial input. They cannot infer correlations between the missing surface and the observed surface.   

Another category~\cite{yuan18pcn, lyne19topnet, dai2017cnncomplete, xie2020grnet, Park_2019_deepsdf, Peng2020convonet, groueix2018atlasnet, xiang21snow} consists of data-driven techniques, which implicitly learn a parametric shape space model. Given an incomplete shape, they find the best reconstruction using the underlying generative model to generate the complete shape. This methodology has enjoyed success for specific categories of models such as faces~\cite{Blanz:1999:Morphable,Zhao2003,Xiong:2013:SDM,DBLP:conf/eccv/RanjanBSB18} and human body shapes~\cite{Anguelov:2005:SCAPE,Loper:2015:SMPL,DBLP:conf/cvpr/PavlakosCGBOTB19,DBLP:conf/iccv/KolotourosPBD19,DBLP:conf/cvpr/KanazawaBJM18}, but they generally cannot recover shape details due to limited training data and difficulty in synthesizing geometric styles that exhibit large topological and geometrical variations. 

This paper introduces a shape completion approach that combines the strengths of the two categories of approaches described above. Although it remains difficult to capture the space of geometric details, existing approaches can learn high-level compositional rules such as spatial correlations of geometric primitives and parts among both the observed and missing regions. We propose to leverage this property to guide similar region retrieval and fusion on a given shape for geometry completion.

Specifically, given an input incomplete shape, the proposed approach first predicts a coarse completion using an off-the-shelf method. The coarse completion does not necessarily capture the shape details but it provides guidance on locations of the missing patches. For each coarse voxel patch, we learn a shape distance function to retrieve top-$k$ detailed shape patches in the input shape. The final stage of our approach learns a deformation for each retrieved patch and a blending function to integrate the retrieved patches into a continuous surface. The deformation prioritizes the compatibility scores between adjacent patches. The blending functions optimize the contribution of each patch and ensure surface smoothness. 

Experimental results on the ShapeNet dataset~\cite{shapenet2015} show that our approach outperforms existing shape completion techniques for reconstructing shape details both qualitatively and quantitatively. 

In summary, our contributions are: 
\begin{itemize}
    \item We propose a non-parametric shape completion method based on patch retrieval and deformation. 
    \item Our method preserves local shape details while enforcing global consistency.
    \item Our method achieves state-of-the-art shape completion results compared with various baselines. 
\end{itemize}

%% file: 02_related.tex
\section{Related Work}

\begin{figure*}[t!]
\centering
\begin{overpic}[width=\textwidth]{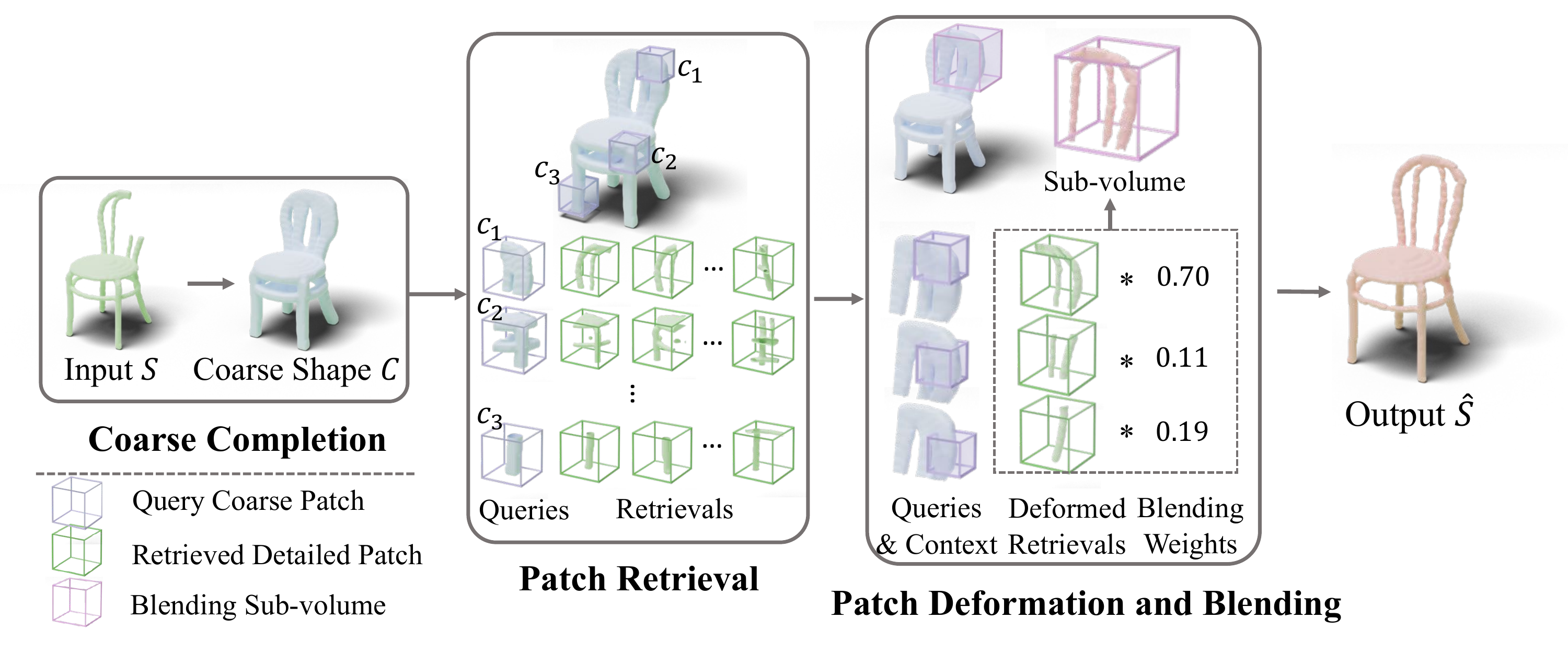}
\end{overpic}
\caption{
\small{Approach pipeline. Given an incomplete shape $S$, we first predict a coarse shape $C$ with the rough structure and no details. For each patch on $C$, $K$ detailed patch candidates are retrieved from the input shape. Then we predict deformations and blending weights for all retrieved candidates. Finally, the output shape $\hat{S}$ is computed by summing up the deformed patches and their blending weights.}
}

\label{Figure:pipeline}
\end{figure*}

\noindent \textbf{Shape Completion.}
Shape completion is a crucial and long-studied task in geometry processing. Non-data-driven works ~\cite{Sorkine04least, Nealen06laplacian, Kazhdan06poisson, Kazhdan13poisson} address hole-filling in a purely geometric manner. Without any high-level prior on the resulting shape, they target filling holes in a ``smooth-as-possible'' manner with membranes.  To complete more complex shapes, several works~\cite{sung15datadriven, Kim123dinddor, Li15database,pauly05example, Li16rgbd, Rock15completing} rely on data-driven methods to get the structure priors or part references. Similarly to our method, \cite{pauly05example, Li16rgbd, Rock15completing} retrieve some candidate models from a database, then perform a non-rigid surface alignment to deform the retrieved shape to fit the input. However, our approach operates at the patch level and can reconstruct shapes that are topologically different from those in the training data.   
With the development of deep learning, neural networks can be used to learn a shape prior from a large dataset and then complete shapes. Voxel-based methods ~\cite{wu20163dgan, dai2017cnncomplete, wu15shapenets} are good at capturing rough structures, but  are limited to low resolution by the cubic scaling of voxel counts. Our framework is especially designed to circumvent these resolution limitations.
Alternatively, point cloud completion~\cite{yuan18pcn, lyne19topnet, xie2020grnet, pan2021vrcnet, yu2021pointr, Xie21sparenet,  wang2021voxel, xiang21snow} has become a popular venue as well. 
\cite{xie2020grnet, huang20pfnet, Wang_2020_cascade} use coarse-to-fine structures to densify the output point cloud and refine local regions. NSFA~\cite{zhang20detail} and HRSC~\cite{han17HRSC} used a two stage method to infer global structures and refine local geometries.  SnowflakeNet~\cite{xiang21snow} modeled the progressive generation hierarchically and arranged the points in locally structured patterns.
As point clouds are sparse and unstructured, it is difficult to recover fine-grained shape details.
3D-EPN\cite{dai2017cnncomplete} and our method both use coarse-to-ﬁne and patch-based pipelines. However, their method only retrieves shapes from the training set and directly copies the nearest patches based on low-level concatenation of distance fields. Our method retrieves patch-level details from the input and jointly learns deformation and blending, which enables our method to handle complex details as well as maintain global coherence.

\noindent \textbf{Patch-based Image In-painting.}
In the 2D domain, many works utilize detailed patches to get high-resolution image inpainting results.
Traditional methods\cite{Efros99texure, patchmatch, kwatra:2003:graphcut, Hays:2007million} often synthesize textures for missing areas or expanding the current images. 
PatchMatch\cite{patchmatch} proposed an image editing method by efficiently searching and replacing local patches. 
SceneComp\cite{Hays:2007million} patched up holes in images by finding similar image regions in a large database. 
Recently, with the power of neural networks,  more methods\cite{tseng20RetrieveGAN, li2019pastegan, qi18semiparam, yang2020learning, ren2019structureflow, Yu2021memaug} use  patch-guided generation to get finer details.
\cite{tseng20RetrieveGAN, li2019pastegan, qi18semiparam} modeled images to scene graphs or semantic layouts and retrieve image patches for each graph/layout component. 
\cite{yang2020learning, ren2019structureflow, Yu2021memaug} add transformers~\cite{NIPS2017transformer},  pixel flow and patch blending to get better generation results respectively.  
Our method leverages many insights from the 2D domain, however these cannot be directly transferred to 3D, for two reasons: i) the signals are inherently different, as 2D pixels are spatially-dense and continuous, while voxels are sparse and effectively binary; ii) the number of voxels in a domain scales cubically with resolution, as opposed to the quadratic scaling of pixels. This significantly limits the performance of various algorithms. The novel pipeline proposed herein is tailor-made to address these challenges.


\noindent \textbf{3D Shape Detailization.}
Adding or preserving details on 3D shapes is an important yet challenging problem in 3D synthesis. Details can be added to a given surface via a reference 3D texture~\cite{Takayama11GeoBrush, Hertz2020texure, kun06quilting}. More relevant to use various geometric representations to synthesize geometric details~\cite{chen2021decor, Chabra20DLS, Genova19localdeep, Chen21Multiresolution, d2im_2021_cvpr}.  DLS~\cite{Chabra20DLS} and LDIF~\cite{Genova19localdeep} divide a shape to different local regions and reconstruct local implicit surfaces.  D2IM-Net~\cite{d2im_2021_cvpr} disentangles shape structure and surface details. DECOR-GAN~\cite{chen2021decor} trained a patch-GAN to transfer details from one shape to another. In our case, we focus on the task of partial-to-full reconstruction, and use detailization as a submodule during the process.

\noindent \textbf{3D Generation by Retrieval.}
Instead of synthesizing shapes from scratch with a statistical model, it is often effective to simply retrieve the nearest shape from a database\cite{tatarchenko2019svr,Kuo21retriv1, Mathias12retriv1,Tangelder04retriv1}. This produces high-quality results at the cost of generalization. Deformation-aware retrieval techniques\cite{Uy20retriv2, Uy21retriv2, Schulz17retriv2,Nan12retriv2} improve the representation power from a limited database. Our method also combines deformation with retrieval, but our retrieval is at the level of local patches from the input shape itself. 
RetrievalFuse\cite{Siddiqui21RetrievalFuse} retrieves  patches from a database for scene reconstruction. An attention-based mechanism is then used to regenerate the scene, guided by the patches. In contrast, we directly copy and deform retrieved patches to fit the output, preserving their original details and fidelity. 

%% file: 03_overview.tex
\section{Overview}
\label{sec::overview}

Our framework receives an incomplete or a partial shape $S$ as input and completes it into a full \emph{detailed} shape $\hat{S}$. Our main observation is that local shape details often repeat and are consistent across different regions of the shape, up to an approximately rigid deformation. Thus, our approach extracts local regions, which we call \emph{patches}, from the given incomplete shape $S$, and uses them to complete and output a full complete shape. 
In order to analyze and synthesize topologically diverse data using  convolutional architectures, we represent shapes and patches as voxel grids with occupancy values, at a resolution of $s_\text{shape}$ cells.

The key challenges facing us are choosing patches from the partial input, and devising a method to deform and blend them into a seamless, complete detailed output. This naturally leads to a three-stage pipeline: (i) 
complete the partial input to get a coarse complete structure $C$ to guide detail completion; (ii) for each completed coarse patch in $C$, retrieve candidate detailed patches from the input shape $S$;  (iii) deform and blend the retrieved detailed patches to output the complete detailed shape $\hat{S}$ (see Figure~\ref{Figure:pipeline}). 
Following is an overview of the process; we elaborate on each step in the following sections.

\noindent \textbf{Coarse Completion.}
We generate a full coarse shape $C$ from the partial input $S$ using a simple 3D-CNN architecture. Our goal is to leverage advances in 3D shape completion, which can provide coarse approximations of the underlying ground truth, but does not accurately reconstruct local geometric details.

\noindent \textbf{Patch Retrieval (Section \ref{sec::retrival}).} 
We train another neural network to retrieve $k$ candidate detailed patches from $S$ for each coarse patch in $C$. Namely, we learn geometric similarity, defined by a rigid-transformation-invariant distance $d$, between the coarse and detailed patches.

\noindent \textbf{Deformation and Blending of Patches (Section \ref{sec::deform}).} 
Given $k$ candidate patches, we use a third neural network to predict rigid transformations and blending weights for each candidate patch, which together define a deformation and blending of patches for globally-consistent, plausible shape completion.

%% file: 04_1Retrieval.tex
\begin{figure}[t!]
\centering
\begin{overpic}[width=0.9\textwidth]{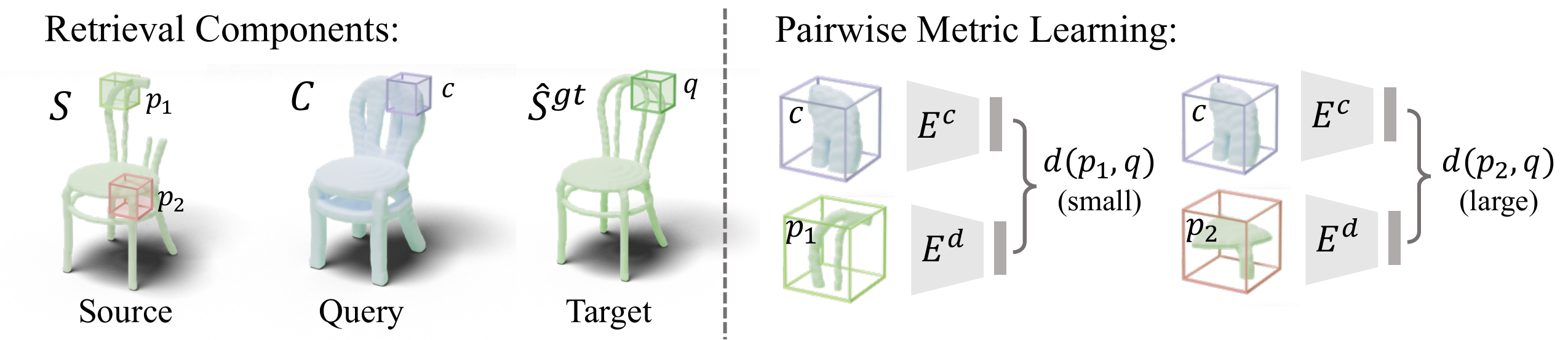}
\end{overpic}
\caption{
Retrieval learning. We learn a feature mapping to predict geometric distances between the query coarse patches and the sampled detailed patches. We use the geometric distances between the GT detailed patches and the sampled patches as the supervision.
Distances for patches that are close up to a rigid transformation are small. Otherwise, distances are large. 
}
\label{Figure:retrieval_learning}
\end{figure}

\section{Patch Retrieval}
\label{sec::retrival}
The input to this stage is the partial detailed shape $S$, and a coarse and completed version of the shape, $C$. The goal of this step is to retrieve a set of patch candidates that can be deformed and stitched to get a fully detailed shape. A \emph{patch} is a cube-shaped sub-region extracted from a shape, composed of $s_\text{patch}^3$ voxels. Our patch sampling process $\mathcal{P}$ takes a shape as input and outputs a collection of patches, where coarse patches $\mathcal{P}(C)$ serve as queries and detailed patches from the partial input $\mathcal{P}(S)$ as sources. 

In order to decide whether a retrieved detailed patch could be an apt substitution for a true detailed patch, we propose a \emph{geometric distance} metric invariant to rigid deformations (Section~\ref{sec::re::geometry}). This geometric distance will be used to supervise the neural network used during testing time, which learns similarities between coarse patches $\mathcal{P}(C)$ and their detailed counterparts $\mathcal{P}(S)$ (Section~\ref{sec::re::learning}). Finally, we describe how to use this network at inference time to retrieve candidate patches for the full shape (Section~\ref{sec::re::full}).

\subsection{Geometric Distance}
\label{sec::re::geometry}
We define a measure of geometric distance, $d$, between two \emph{detailed} shape patches $(p_1, p_2)$. This metric should be agnostic to their poses, since the pose can be fixed at the deformation stage, hence we define the distance as the minimum over all possible rigid transformations of the patch:
\begin{equation}
    d(p_1,p_2) = \min_{T} \frac{\|T(p_1)-p_2\|_1}{\|T(p_1)\|_1 + \|p_2\|_1}
    \label{Eq:Geometric:Distance}
\end{equation}
where $T$ is a rigid motion composed with a possible reflection, i.e., $T=(R, \mathbf{t}, f)$,  $R\in SO(3)$ is the rotation, $\mathbf{t}\in \mathbb{R}^3$ is the translation, $f \in \{0,1\}$ denotes if a reflection is applied, and $||\cdot||_1$ denotes the L1 norm of positional vectors to patch centers. To practically compute this distance, we run ICP~\cite{DBLP:journals/pami/BeslM92}, initialized from two transformations (with and without reflection enabled) that align patch centers.
While this geometric distance can be computed for detailed patches, at inference time we only have coarse patches. Therefore, we train a network to embed coarse patches into a latent space in which Euclidean distances match the geometric distances of the detailed patches they represent. 

\subsection{Metric embedding}
\label{sec::re::learning}
We train two neural networks to act as encoders, one for coarse patches and one for detailed patches, $E^\text{c}$ and $E^\text{d}$, respectively. We aim to have the Euclidean distances between their generated codes reflect the distances between the true detailed patches observed during training. Given a coarse patch $c \in \mathcal{P}(C)$ with its true corresponding detailed patch $q \in \mathcal{P}(\hat S^\text{gt})$, as well as a some other detailed patch $p \in \mathcal{P}(S)$, we define a metric embedding loss:
\begin{equation}
    L_\text{r} = \sum_{(c, p, q) \in \mathcal{T}}  \| \|E^\text{c}(c)-E^\text{d}(p)\|_2 - d(p, q)) \|_2  .
\end{equation}
where $ d(p, q)$ is the geometric distance defined in Equation (\ref{Eq:Geometric:Distance}).
Our training triplets are composed of true matches and random patches: $\mathcal{T}=\mathcal{T}_\text{true} \cup \mathcal{T}_\text{rnd}$. Where in both sets $c$ is a random coarse patch, $q$ is the corresponding true detailed patch. We either set $p=q$ for  $\mathcal{T}_\text{true}$ or randomly sample $p \in \mathcal{P}(S)$ for $\mathcal{T}_\text{rnd}$. See Figure~\ref{Figure:retrieval_learning} for an illustration.

\subsection{Retrieval on a Full Shape}
\label{sec::re::full}
We can now use trained encoder networks at inference time to retrieve detailed patches for each coarse patch. First, we encode all the detailed patches in $\mathcal{P}(S)$ via $E^\text{d}$. Similarly, for each non-empty coarse patch $c \in \mathcal{P}(C)$ with lowest corner at location $l$, we encode it with $E^\text{c}$ and find the $K$-nearest-neighbor detailed codes. We store the list of retrieved patches for each location, denoted as $\mathcal{R}_l$. 

We sample the coarse patches using a fixed-size ($s_\text{patch}^3$) sliding window with a stride $\gamma_\text{patch}$. 
Note that in the retrieval stage we do not assume that we know which parts of the detailed shape need to be completed. Since our feature learning step observed a lot of positive coarse/fine pairs with the detailed input, we found that the input is naturally reconstructed from the retrieved detailed patches.

%% file: 04_2Deformation.tex
\section{Deformation and Blending of Patches}
\label{sec::deform}

\begin{table*}[t!]
\centering
 \begin{tabular}{ c |c | c c c c c c c c} 
 \hline
   & Average & Chair  & Plane & Car & Table & Cabinet & Lamp & Boat & Couch  \\
 \hline
 
 AtlasNet\cite{groueix2018atlasnet} & 7.03 & 6.08  &  2.32 & 5.32 & 5.38 & 8.46 & 14.20 & 6.01 & 8.47 \\ 
 Conv-ONet\cite{Peng2020convonet} & 6.42 & 2.91   & 2.29 & 8.60 & 7.94 &12.6  & 5.82  & 4.03 & 7.21\\
TopNet\cite{lyne19topnet} & 6.30 & 5.94 & 2.18 & 4.85 & 5.63 & 5.13 & 15.32 & 5.60 & 5.73 \\
3D-GAN\cite{wu20163dgan} & 6.00 & 6.02 & 1.77 & 3.46 & 5.08 & 7.29 & 12.23 & 7.20 & 4.92 \\
PCN\cite{yuan18pcn} & 4.47 & 3.75 & 1.45 & 3.58 & 3.32 & 4.82 & 10.56  & 4.22 & 4.03\\ 
 GRNet\cite{xie2020grnet} & 2.69 & 3.27  & 1.47 & 3.15 & 2.43 & 3.35 & 2.54 & 2.50 & 2.84\\
 VRCNet\cite{pan2021vrcnet} & 2.63 & 2.96  & 1.30 & 3.25 & 2.35 & 2.98 & 2.86 & 2.23 & 3.13 \\
 SnowflakeNet\cite{xiang21snow} & 2.06 & 2.45 & \textbf{0.72} & 2.55 & 2.15 & 2.76 & 2.17 & 1.33 & 2.35 \\
 \hline
 PatchRD (Ours) & \textbf{1.22 }& \textbf{1.08}  & 0.98 & \textbf{1.01} & \textbf{1.32} & \textbf{1.45} & \textbf{1.23} & \textbf{0.99} & \textbf{1.67} \\
 \hline
\end{tabular}
\caption{Shape completion results on the random-crop dataset on 8 ShapeNet categories. We show the $L_2$ Chamfer distance (CD) $(\times 10^3)$ between the output shape and the ground truth 16384 points from PCN dataset\cite{yuan18pcn} (lower is better). Our method reduces the CD drastically compared with the baselines. 
}
\label{Table::CD}
\end{table*}

The input to this stage is the coarse shape $C$, partial input $S$, and the retrieval candidates. The output is the full detailed shape $\hat S$, produced by deforming and blending the retrieved patches. As illustrated by Figure~\ref{Figure:pipeline} we first apply a rigid transformation to each retrieved patch and then blend these transformed patches into the final shape. 
Our guiding principle is the notion of partition-of-unity~\cite{Ohtake:2003:MLP}, which blends candidate patches with optimized transformations into a smooth completion. Unlike using fixed weighting functions, we propose to learn the blending weights. These weights serve the role of selecting candidate patches and stitching them smoothly. 

We observe that learning the blending weights requires some context (our method needs to be aware of at least a few neighboring patches), but does not require understanding the whole shape (coarse shape and retrieved patches already constrain the global structure). Thus, to maximize efficiency and generalizability, we opt to perform deformation and blending at the meso-scale of subvolumes $V \subset S$ with size $s_\text{subv}$.

Next, we provide more details on our blending operator (Section~\ref{Sec::operator}) and how to learn it from the data (Section~\ref{sec::learning-deformation-blending}). 

\subsection{The Deformation and Blending Operator}
\label{Sec::operator}

Given a subvolume $V$, we first identify $[r_m, m=1...M]$ an ordered list of $M$ best patches to be considered for blending. These patches are from the retrieved candidates $\mathcal{R}_l$ such that $l\in V$, and sorted according to two criteria: (i) retrieval index, (ii) $x,y,z$ ordering. If more than $M$ such patches exist, we simply take the first $M$. Each patch $r_m$ is transformed with a rigid motion and possible reflection: $T_m$, and we have a blending weight for each patch at every point $x$ in our volume: $\omega_m[x]$. The output at voxel $x$ is the weighted sum of the deformed blending candidates:

\begin{equation}
     V[x] = \frac{1}{\xi[x]}\sum_{m=1...M} \omega_m[x] \cdot T_m(r_m) [x]
     \label{Eq:Blending}
\end{equation}
where $\omega_m[x]$ is the blending weight for patch $m$ at voxel $x$, and $T_m(r_m)$ is the transformed patch (placed in the volume $V$, and padded with $0$), and $\xi[x]=\sum_{m=1..M}\omega_m[x]$ is the normalization factor.
At inference time, when we need to reconstruct the entire shape, we sample $V$ over the entire domain $\hat S$ (with stride $\gamma_\text{subv}$), and average values in the region of overlap.

\subsection{Learning Deformation and Blending}
\label{sec::learning-deformation-blending}
Directly optimizing deformation and blending is prone to being stuck in local optimum. To address this we develop a neural network to predict deformations and blending weights and train it with reconstruction and smoothness losses. 


\noindent\textbf{Prediction network.} We train a neural network $g$ to predict deformation and blending weights. The network consists of three convolutional encoders, one for each voxel grid: the coarse shape (with a binary mask for the cropped subvolume $V$), the partial input, and the tensor of retrieved patches ($M$ channels at resolution of $V$). 
We use fully-connected layers to mix the output of convolutional encoders into a bottleneck, which is than decoded into deformation $T$ and blending $\omega$ parameters. 


\noindent \textbf{Reconstruction loss.} The first loss
$L_\text{rec}$ aims to recover the target shape $\hat S^\text{gt}$:
\begin{equation}
    L_\text{rec} = \| V^\text{gt} - V\|_2,
\end{equation}
where $V^\text{gt} \subset \hat S^\text{gt} $ and $V \subset \hat S$ are corresponding true and predicted subvolumes (we sample $V$ randomly for training).


\noindent \textbf{Blending smoothness loss.} The second loss $L_\text{sm}$ regularizes patch pairs. Specifically, if two patches have large blending weights for a voxel, then their transformations are forced to be compatible on that voxel: 
$$
    L_\text{sm} = \sum_{x \in \mathcal{V}} \sum_{m,n} \|\omega_m[x] \cdot \omega_n[x] \cdot (T_m(r_m)[x] - T_n(r_n)[x]) \|
$$

Where $x$ iterates over the volume and $m,n$ over all retrieved patches. Note that $r_m$ and $r_n$ are only defined on a small region based on where the patch is placed, so this sum only matters in regions where transformed patches $T_m(r_m)$ and $T_n(r_n)$ map to nearby points $x$ accordingly.

\noindent \textbf{Final Loss}
The final loss term is 
\begin{equation}
    L = L_\text{rec} + \alpha L_\text{sm}.
\label{eq:blending_loss}
\end{equation}

%% file: 05_results.tex
\section{Experiments}

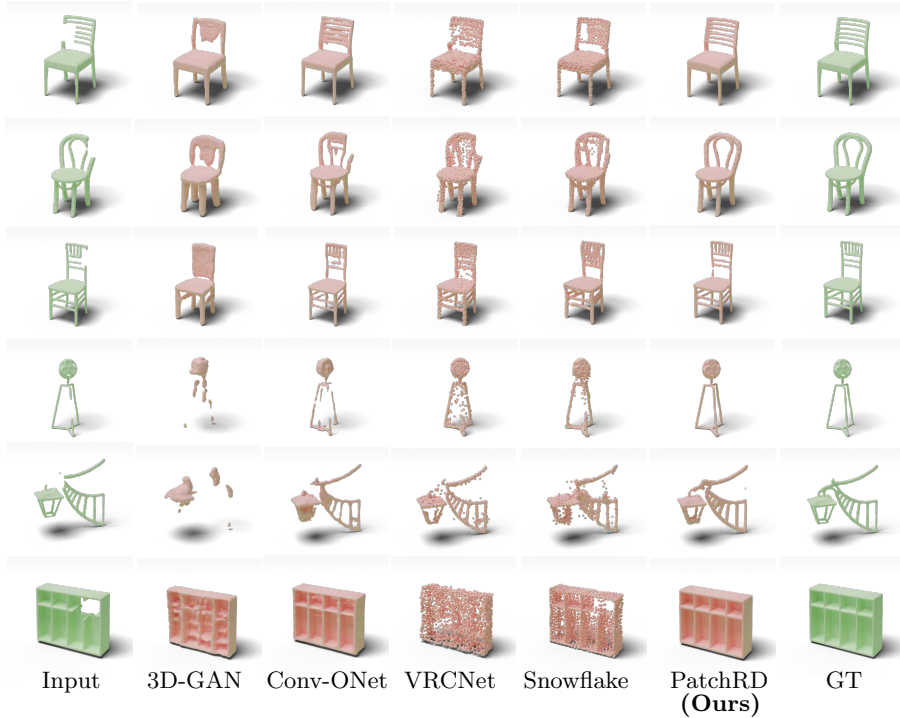
\begin{figure*}[t!]
\centering
\begin{overpic}[width=1\textwidth]{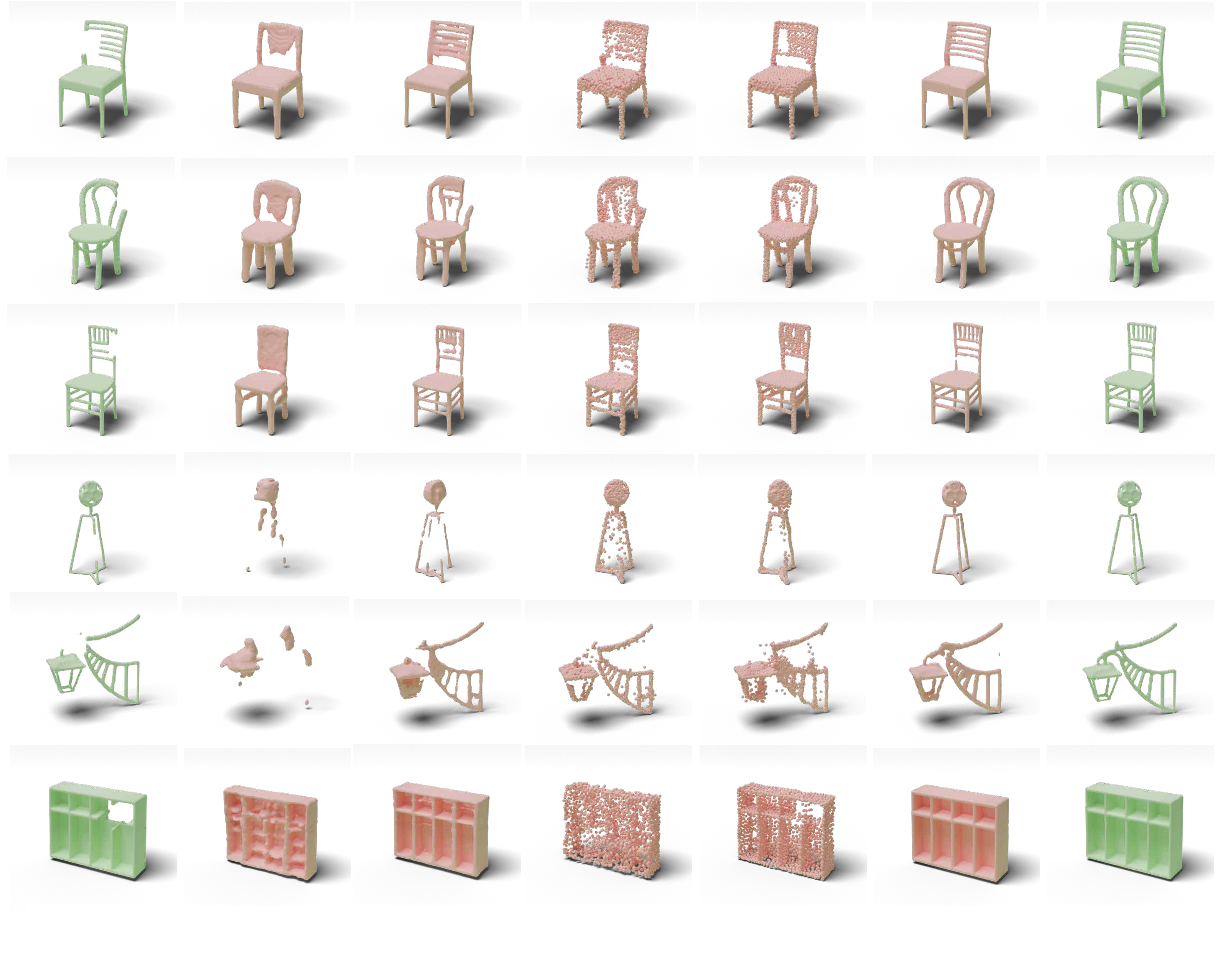}
\put(4.5,2.5) {Input}
\put(16,2.5) {3D-GAN} 
\put(29,2.5) {Conv-ONet} 
\put(44,2.5) {VRCNet} 
\put(57,2.5) {Snowflake} 
\put(73,2.5) {PatchRD}
\put(74,0) {\textbf{(Ours)}}
\put(89,2.5) { GT}

\end{overpic}
\caption{Qualitative shape completion results on the Random-Crop Dataset.  Our results recover more geometric details and keep the shape smooth  while other baselines often produce coarse, noisy, or discontinuous results.  } 

\label{Figure:Inpaint:randcrop}
\end{figure*}

We primarily evaluate our method on the detail-preserving shape completion benchmark (Section~\ref{sec:setup}), and demonstrate that our method outperforms state-of-the-art baselines (Section~\ref{main_results}). We further demonstrate that our method can generalize beyond the benchmark setup, handling real scans, data with large missing areas, and novel categories of shapes (Section~\ref{sec::results::otherapp}). Finally, we run an ablation study (Section~\ref{sec::results::ablation}) and evaluate sensitivity to the size of the missing area (Section~\ref{sec::results::cropratio}).

\subsection{Experimental Setup} 
\label{sec:setup}

\noindent \textbf{Implementation Details.} 
We use the following parameters for all experiments. The sizes of various voxel grids are: $s_\text{shape}=128, s_\text{patch}=18, s_\text{subv}=40$ with strides $\gamma_\text{patch}=4, \gamma_\text{subv}=32$. 
We sample $|\mathcal{T}_\text{rnd}|=800$ and $|\mathcal{T}_\text{true}|=400$ triplets to train our patch similarity (Section~\ref{sec::re::learning}). Our blending operator uses $M=400$ best retrieved patches (Section~\ref{Sec::operator}).
We set $\alpha=10$ for Equation~\ref{eq:blending_loss}. To improve performance we also define our blending weights $\omega_m$ at a coarser level than $V$. In particular, we use windows of size $s_\text{blend}^3=8^3$ to have constant weight, and compute the blending smoothness term at the boundary of these windows.

\noindent \textbf{Dataset.} We use shapes from ShapeNet\cite{shapenet2015}, a public large-scale repository of 3D meshes to create the completion benchmark. 
We pick eight shape categories selected in prior work PCN~\cite{yuan18pcn}. For each category, we use the same subset of training and testing shapes with 80\% / 20\% split as in DECOR-GAN work~\cite{chen2021decor}. For voxel-based methods, we convert each mesh to a $128^3$ voxel grid, and for point-based baselines, we use the existing point clouds with 16384 points per mesh~\cite{yuan18pcn}. We create synthetic incomplete shapes by cropping (deleting) a random cuboid with $10\%-30\%$ volume with respect to the full shape. 
This randomly cropped dataset is generated to simulate smaller-scale data corruption.  
We also show results on planar cutting and point scans in Section~\ref{sec::results::otherapp}. 

\begin{table}[t!]
\centering
\begin{adjustbox}{width=0.5\columnwidth,center}
\begin{tabular}{ c | c c c } 
 \hline
  & Conv-ONet\cite{Peng2020convonet} & 3D-GAN\cite{wu20163dgan} & PatchRD(Ours) \\
  \hline 
 FID &174.72 &  157.19 & \textbf{11.89}\\
 \hline
\end{tabular}
\end{adjustbox}
\caption{FID comparison on the chair class (note that we can only apply this metric to volumetric baselines). Our method produces more plausible shapes.
}
\label{Table::FID}
\end{table}

\noindent \textbf{Metrics.} 
To evaluate the quality of the completion, we use the $L_2$ Chamfer Distance (CD) with respect to the ground truth detailed shape. Since CD does not really evaluate the quality of finer details, we also use Frechet Inception Distance (FID), to evaluate plausibility. FID metric computes the distance of the layer activations from a pre-trained shape classifier. We use 3D VGG16\cite{Simonyan15vgg}) trained on ShapeNet and activations of the first fully connected layer.

\noindent \textbf{Baseline Approaches}
To the best of our knowledge, we are the first to do the 3D shape completion task on the random-cropped dataset. Considering the task similarity, we compare our method with the other shape completion and reconstruction baselines. 

Our baselines span different shape representations: PCN\cite{yuan18pcn}, TopNet\cite{lyne19topnet}, GRNet\cite{xie2020grnet}, VRCNet\cite{pan2021vrcnet}, and SnowFlakeNet\cite{xiang21snow} are point-based scan completion baselines,  3D-GAN\cite{wu20163dgan} is a voxel-based shape generation method, Conv-ONet\cite{Peng2020convonet} is an implicit surfaces-based shape reconstruction methods, and AtlasNet\cite{groueix2018atlasnet} is an atlas-based shape reconstruction method. We show our method outperforms these baselines both quantitatively and qualitatively.  

\subsection{Shape Completion Results}
\label{main_results}
Table~\ref{Table::CD} and Table~\ref{Table::FID} show quantitative comparisons between PatchRD and baselines, demonstrating that our method significantly outperforms all baselines. PatchRD achieved superior performance on all categories except airplanes, which shows that it generalizes well across different classes of shapes. 

Specifically, the voxel-based baseline\cite{wu20163dgan} produces coarse shapes where fine details are missing. Point-based baselines\cite{pan2021vrcnet, xiang21snow} often have noisy patterns around on fine structures while our method has clean geometry details. The implicit surface based method\cite{Peng2020convonet} could capture the  details but the geometry is not smooth and the topology is not preserved. Our method keeps the smooth connection between geometry patches. More results can be found in the supplemental materials.
Figure \ref{Figure:Inpaint:randcrop} shows qualitative comparisons. We pick four representative baselines for this visualization including point-based methods that performed the best on the benchmark~\cite{pan2021vrcnet, xiang21snow} as well as voxel-based~\cite{wu20163dgan} and implicit-based methods~\cite{Peng2020convonet}. Our results show better shape quality by recovering local details as well as preserving global shape smoothness.

\begin{figure}[t!]
\centering
\begin{subfigure}{\textwidth}
\begin{overpic}[width=1\textwidth]{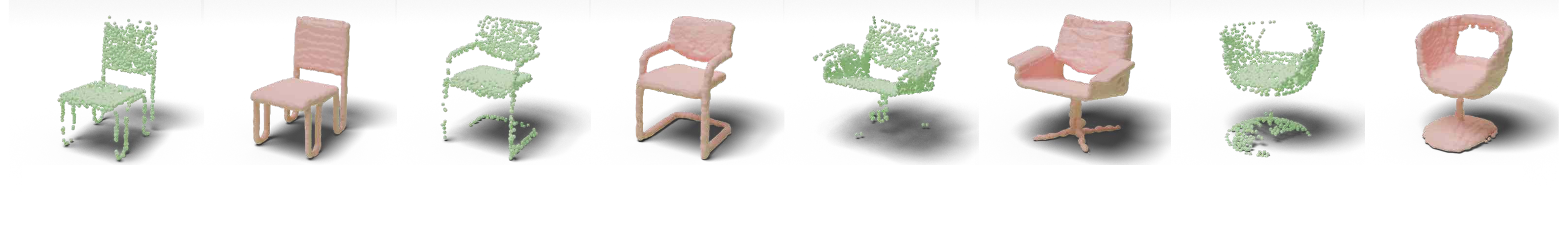}
\put(4,2) {Input }
\put(0,-0.5) {{\scriptsize (Point Cloud)}}
\put(15,2) {Output}
\put(28,2) {Input }
\put(24, -0.5) {{\scriptsize (Point Cloud)}}
\put(39,2) {Output}
\put(53,2) {Input }
\put(49,-0.5) {{\scriptsize (Point Cloud)}}
\put(65,2) {Output}
\put(79,2) {Input}
\put(75, -0.5) {{\scriptsize (Point Cloud)}}
\put(90,2) {Output}
\end{overpic}
\caption{Shape completion results on real scans for ScanNet objects. Our method completes the missing areas and fills the uneven areas with detailed and smooth geometries. }
\label{Figure:real}
\end{subfigure}
\hfill

\begin{subfigure}{1\textwidth}
\begin{overpic}[width=\textwidth]{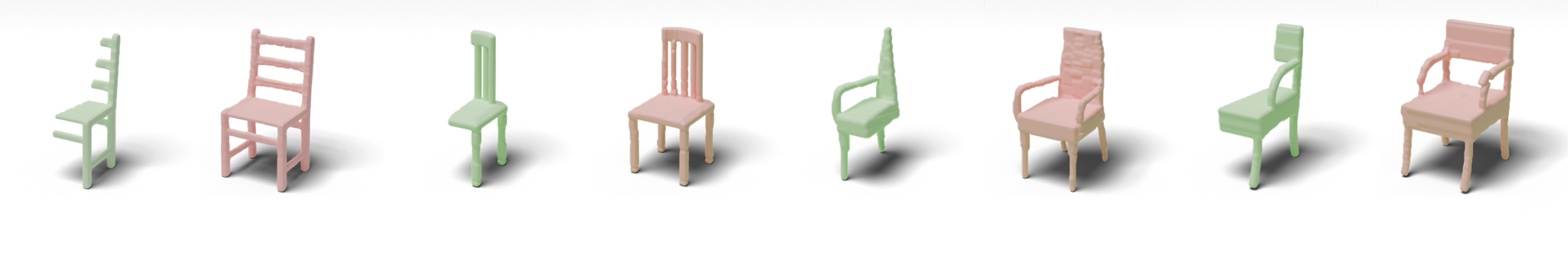}
\put(2.5,1.5) {Input }
\put(13.5,1.5) {Output}
\put(28,1.5) {Input }
\put(39,1.5) {Output}
\put(52,1.5) {Input }
\put(64,1.5) {Output}
\put(77,1.5) {Input}
\put(88,1.5) {Output}
\end{overpic}
\caption{Shape completion results on shapes with large missing areas. Our method recovers geometric details even when given relatively small regions with reference patterns.}
\label{Figure:planecut}
\end{subfigure}
\hfill
\begin{subfigure}{1\textwidth}
  \begin{overpic}[width=1\textwidth]{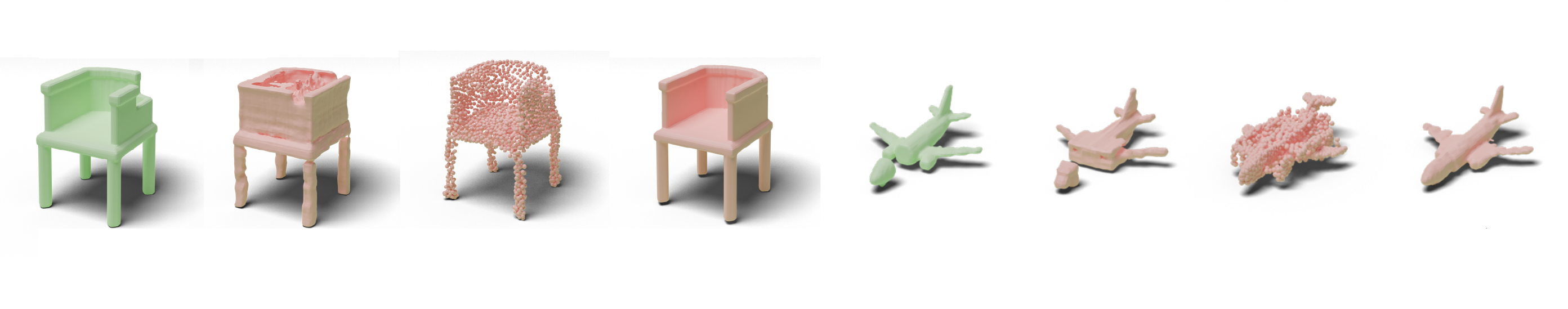}
\put(2.5,17) {Lamp $\rightarrow$ Chair}
\put(2.5,2.5) {Input }
\put(10,2.5) {Conv-ONet\cite{Peng2020convonet}}
\put(28,2.5) {Snow\cite{xiang21snow} }
\put(40,2.5) {PatchRD}
\put(55,17) {Cabinet $\rightarrow$ Plane}
\put(55,2.5) {Input }
\put(62.5,2.5) {Conv-ONet\cite{Peng2020convonet}}
\put(80,2.5) {Snow\cite{xiang21snow}}
\put(90,2.5) {PatchRD}
\end{overpic}
\caption{Testing results on novel categories. We show results trained on lamp and cabinet categories and inferred on lamp  and plane categories, respectively. Our method has better generalization ability. }
\label{Figure:cross}
\end{subfigure}
\caption{More applications on real scans, shapes with large missing areas and novel categories.}
\label{fig:figures}
\end{figure}

\subsection{Other Applications}
\label{sec::results::otherapp}
\noindent \textbf{Real-World Application: Scan Completion}.
We test our method on real-world shape completion from scans. We use shapes from ScanNet\cite{dai2017scannet}, 3D indoor scene dataset as input to our method. Objects in ScanNet  often have some missing parts, especially thin structures,  due to occlusion and incompleteness of scan viewpoints.
We convert these point clouds to voxel grids and apply our completion technique trained on ShapeNet, see results in Figure~\ref{Figure:real}. Note how our method completes the undersampled areas, while preserving the details of the input and smoothly blending new details to the existing content.

\noindent \textbf{Shapes with Large Missing Areas}. We also demonstrate that our method can handle large missing areas (Figure~\ref{Figure:planecut}). In this experiment we cut the shape with a random plane, where in some cases more than half of the shape might be missing. Our method recovers the shape structure and extends the local shape details to the whole shape when only given a small region of reference details. 

\noindent \textbf{Completion on Novel Categories}
We further evaluate the ability of our method to generalize to novel categories. Note that only the prediction of the complete coarse shape relies on any global categorical priors. Unlike other generative techniques that decode the entire shape, our method does not need to learn how to synthesize category-specific details and thus succeeds in detail-preservation as long as the coarse shape is somewhat reasonable. In Figure~\ref{Figure:cross} we demonstrate the output of different methods when tested on a novel category. Note how our method is most successful in conveying overall shape as well as matching the details of the input.

\begin{table*}[t!]
\begin{minipage}{0.5\textwidth}
  \centering
  \includegraphics[width=1\textwidth]{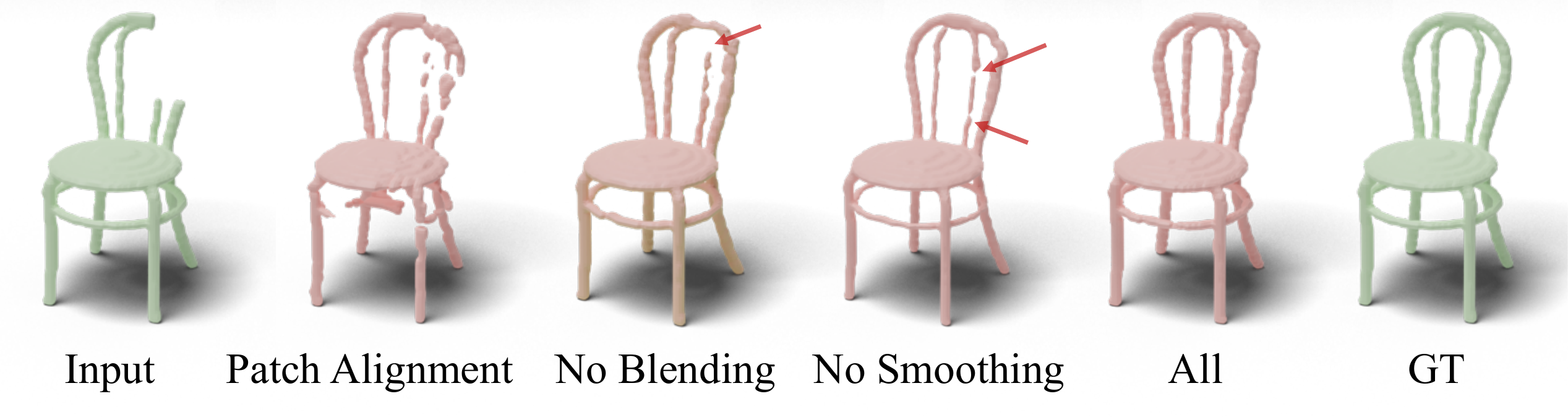}

\end{minipage}
\hfill
\begin{minipage}{0.48\textwidth}
\centering
\begin{tabular}{ c | c c } 
 \hline
  & CD & FID \\
  \hline 
  Patch Alignment & 4.90 & 43.35 \\
  No Blending & 2.03 & 30.25 \\
  No Smoothing & 1.86 & 27.42 \\
  All & \textbf{1.43}& \textbf{11.89} \\
 \hline
\end{tabular}

\end{minipage}
\caption{Ablation study. In the left figure, we visualize the effect of different components in our experiment.  Patch alignment can't get good patch transformation. Results with no blending are subjective to bad retrievals. Results with no smoothing show discontinuity between neighboring patches. Results with all components contain geometric details as well as smoothness.
In the right table, We show the reconstruction error CD and shape plausibility score FID on ShapeNet chair class. Results with all components get both better CD and FID.  }
\label{Table::ablation}
\end{table*}

\subsection{Ablation Study}
\label{sec::results::ablation}

We evaluate the significance of deformation learning, patch blending, and blending smoothness term via an ablation study (see Table~ \ref{Table::ablation}).

\noindent \textbf{No Deformation Learning}
We simply use ICP to align the retrieved patch to the query. Table~\ref{Table::ablation} (Patch Alignment) illustrates that this leads to zigzagging artifacts due to patch misalignments.

\noindent \textbf{No Patch Blending}
Instead of blending several retrieved patches, we simply place the best retrieved patch at the query location. Table \ref{Table::ablation} (No Blending) shows that this single patch might not be sufficient, leading to missing regions in the output.

\noindent \textbf{No Blending Smoothness}
We set the blending smoothness to $L_\text{sm}=0$ to remove our penalty for misalignments at patch boundaries. Doing so leads to artifacts and discontinuities at patch boundaries (Table \ref{Table::ablation}, No Smoothing). 

The quantitative results in Table.\ref{Table::ablation} show that our method with all components performs the best with respect to reconstruction and plausibility metrics.

\subsection{Sensitivity to the Size of Missing Regions}
\label{sec::results::cropratio}
The completion task is often ill-posed, and becomes especially ambiguous as the missing region increases. We evaluate the sensitivity of our method to the size of the region by trying to increase the crop size from 10\% to 50\% of the volume. Table~\ref{Table::crop_ratio} demonstrates that our method can produce plausible completions even under severe crops. However, in some cases it is impossible to reconstruct details that are completely removed. We report quantitative results in Table~\ref{Table::crop_ratio}. While both reconstruction and plausibility error increases for larger crops, we observe that plausibility (FID) score does not deteriorate at the same rate.

\begin{table*}[t!]
\begin{minipage}{0.63\textwidth}
\centering
\includegraphics[width=1\textwidth]{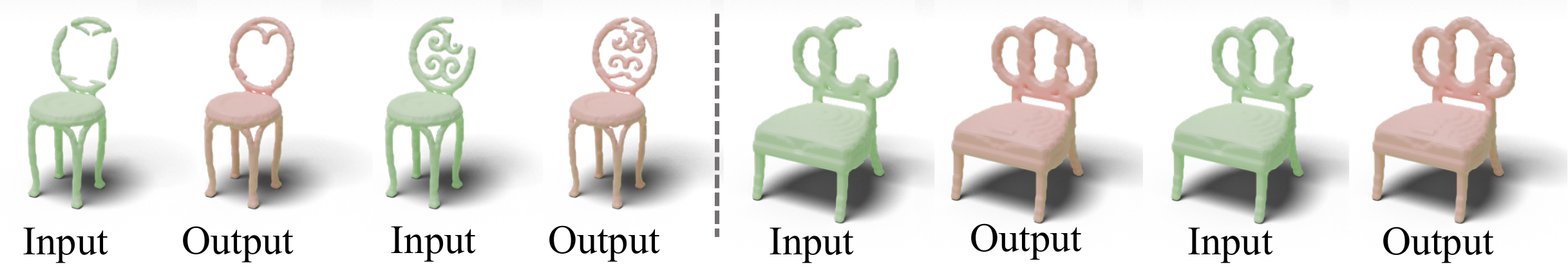}
\end{minipage}
\hfill
\begin{minipage}{0.35\textwidth}
\centering
 \begin{tabular}{ c | c c c  c} 
 \hline
  & 10\% &  20\% &  40\% & 60\% \\
 \hline
 $L_2$-CD & 0.88 & 1.22 & 2.35 & 6.64  \\
 FID & 9.74 & 10.32 & 13.34 & 15.63 \\
 \hline
\end{tabular}
\end{minipage}
\caption{Sensitivity to the size of missing regions. The left figure shows results with different crop ratios. Input geometries and shape contours influence the output shapes. The right table shows the reconstruction error and the shape plausibility with the increase of crop ratios. As the ratio increases, the reconstruction error keeps growing although the output shapes remain fairly plausible.}
\label{Table::crop_ratio}
\end{table*}

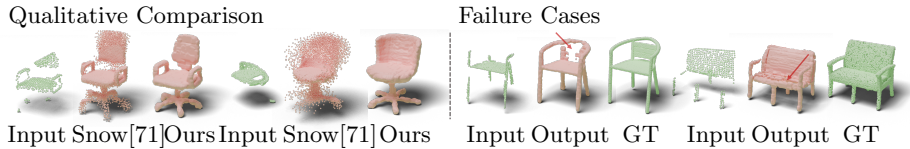
\begin{figure}[t!]
\centering
\begin{overpic}[width=\textwidth]{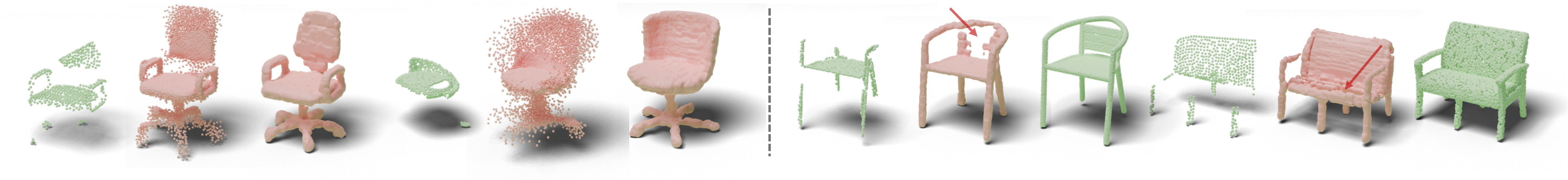}
\put(1,12) {Qualitative Comparison}
\put(0,-1) { Input }
\put(7,-1) { Snow\cite{xiang21snow}}
\put(17,-1) { Ours}
\put(23,-1) { Input}
\put(30,-1) { Snow\cite{xiang21snow}}
\put(40.5,-1) { Ours}
\put(50,12) {Failure Cases}
\put(50,-1) { Input }
\put(57,-1) { Output}
\put(67,-1) { GT}
\put(74,-1) { Input}
\put(81,-1) { Output}
\put(91,-1) { GT}
\end{overpic}
\caption{Qualitative results on the PCN Dataset\cite{yuan18pcn}. On the left, we show our method is able to produce cleaner and more plausible results than the structure-based baseline. On the right, we show some failure cases where shape details are missing in the input shape. }
\label{Figure:failure}
\end{figure}

%% file: 06_conclusions.tex
\section{Conclusions, Limitations and Future Work}

\noindent \textbf{Conclusions.}
This paper proposed a novel non-parametric shape completion method that preserves local geometric details and global shape smoothness.
Our method recovers details by copying detailed patches from the incomplete shape, and achieves smoothness by a novel patch blending term. Our method obtained state-of-the-art completion results compared with various baselines with different 3D representations. It also achieved high-quality results in real-world 3D scans and shapes with large missing areas.

\noindent \textbf{Limitations.}
Our method has two limitations: 
(1) It builds on the assumption that the shape details are present in the partial input shape, which might not hold if large regions are missing in the scan.
For completeness, we still evaluate our method in this scenario using the PCN benchmark, which focuses on large-scale structure recovery from very sparse input. In Figure~\ref{Figure:failure} we show that our method succeeds when there are enough detail references in the input, and fails if the input is too sparse. We also provide quantitative evaluations in supplemental material. These results suggest that our method is better suited for completing denser inputs (e.g., multi-view scans). 

In the future, we plan to address this issue by incorporating patches retrieved from other shapes. 
(2) Our method cannot guarantee to recover the global structure because the retrieval stage is performed at the local patch level. To address this issue, we need to enforce suitable structural priors and develop structure-aware representations.  We leave both for future research.

\noindent \textbf{Future work.}
Recovering geometric details is a hard but important problem. Our method shows that reusing the detailed patches from an incomplete shape is a promising direction. 
In the direction of the patch-based shape completion, potential future work includes: 
(1) Applying patch retrieval and deformation on other 3D representations such as point cloud and implicit surfaces. This can handle the resolution limitation and computation burden caused by the volumetric representation. 
(2) Unifying parametric synthesis and patch-based non-parametric synthesis to augment geometric details that are not present in the partial input shape.

%% file: 07_supp.tex
\section{More Results}
\begin{figure*}
\centering
\begin{overpic}[width=0.9\textwidth]{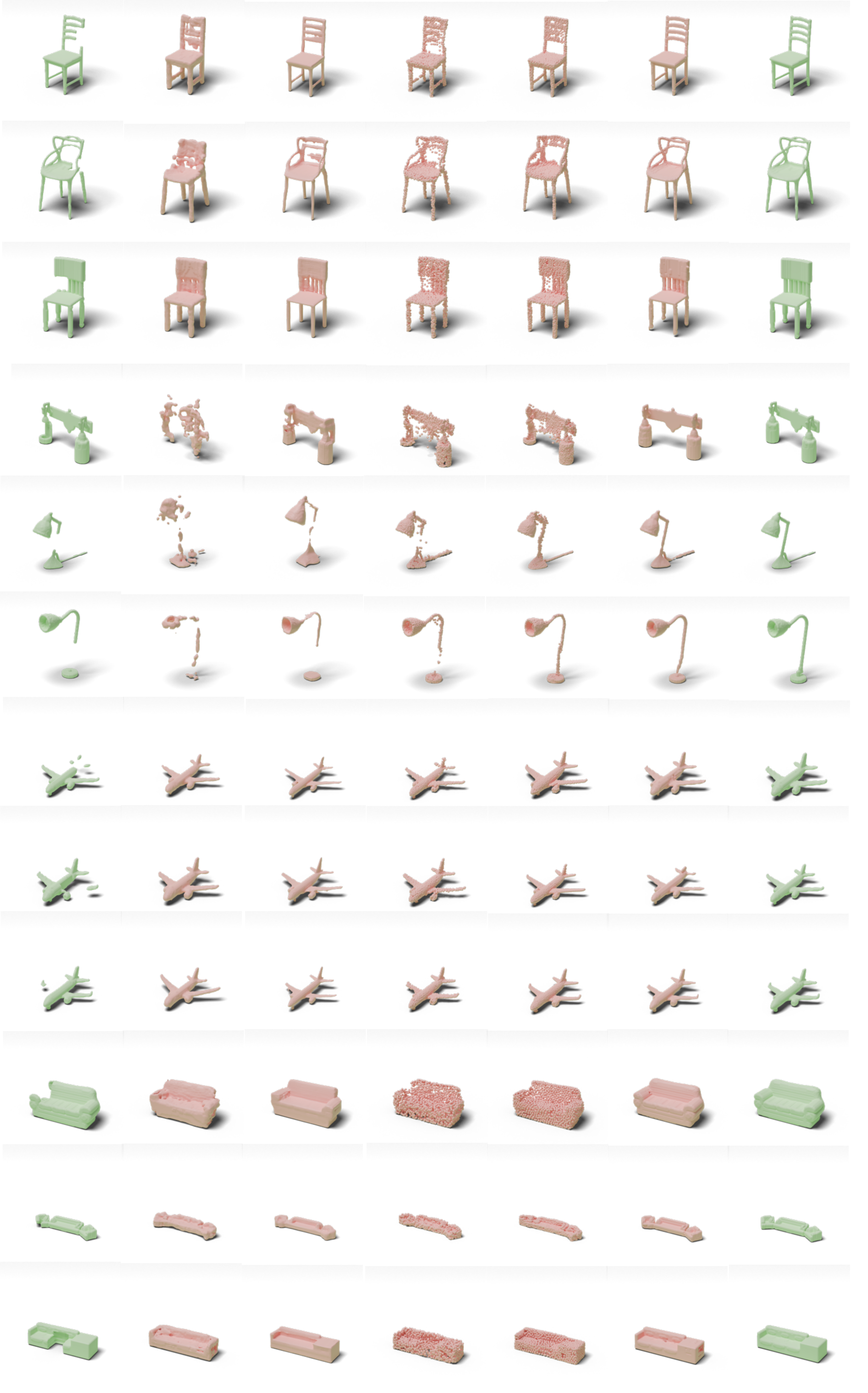}
\put(2,0) {  Input}
\put(9,0) { 3D-GAN }
\put(17,0) { Conv-ONet}
\put(27,0) { VRCNet }
\put(35,0) { Snowflake}
\put(44.5,0) { PatchRD}
\put(45.5,-1.5) {\textbf{(Ours)}}
\put(54.5,0) { GT}
\end{overpic}
\caption{More qualitative shape completion results on the Random-Crop Dataset. }
\label{Figure:supp_crop1}
\end{figure*}

\begin{figure*}
\centering
\begin{overpic}[width=0.9\textwidth]{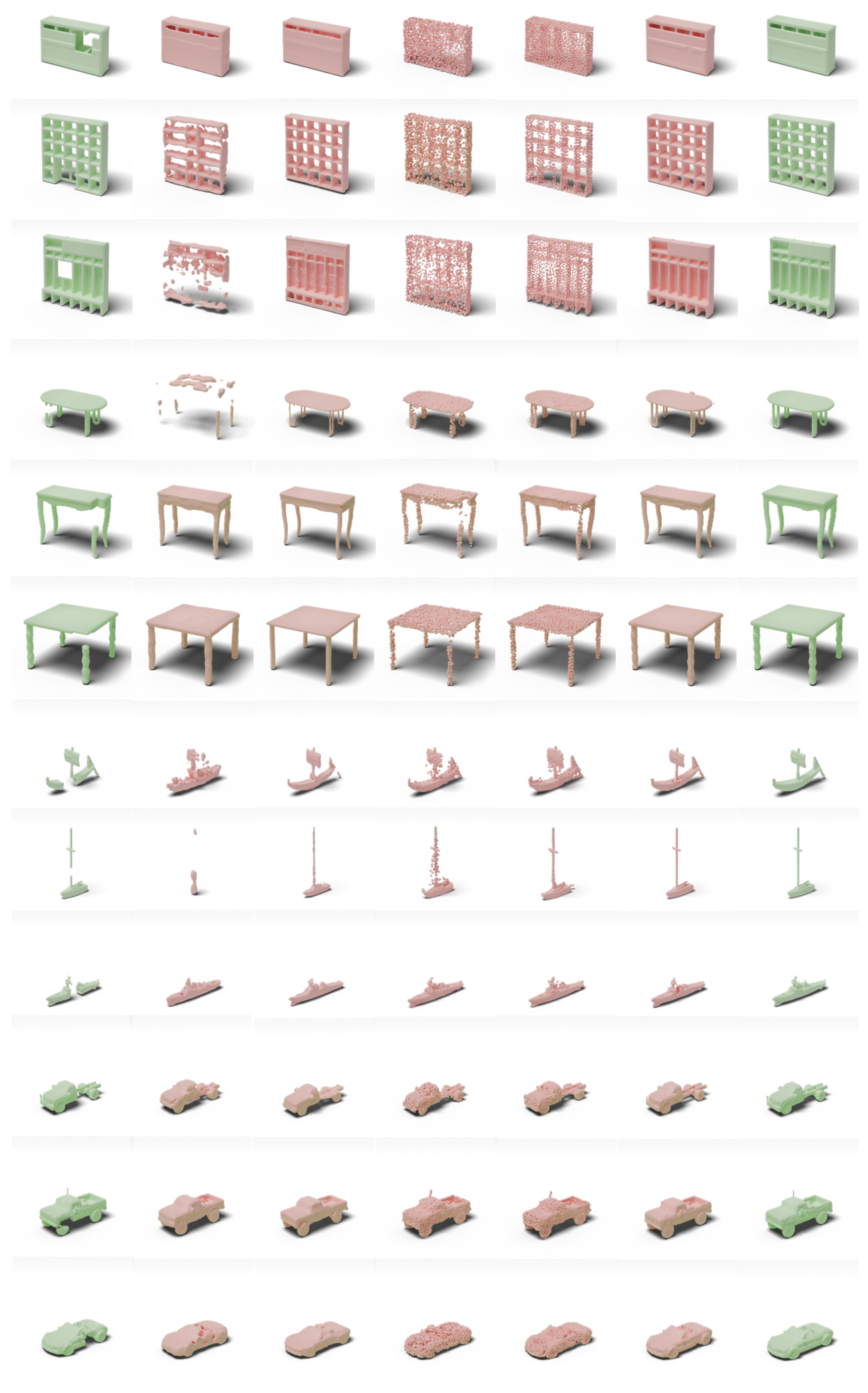}
\put(2,0) {  Input}
\put(9,0) { 3D-GAN }
\put(17,0) { Conv-ONet}
\put(27,0) { VRCNet }
\put(35,0) { Snowflake}
\put(44.5,0) { PatchRD}
\put(45.5,-1.5) {\textbf{(Ours)}}
\put(54.5,0) { GT}
\end{overpic}
\caption{More qualitative shape completion results on the Random-Crop Dataset.}
\label{Figure:supp_crop2}
\end{figure*}

\begin{figure*}
\centering
\begin{overpic}[width=0.9\textwidth]{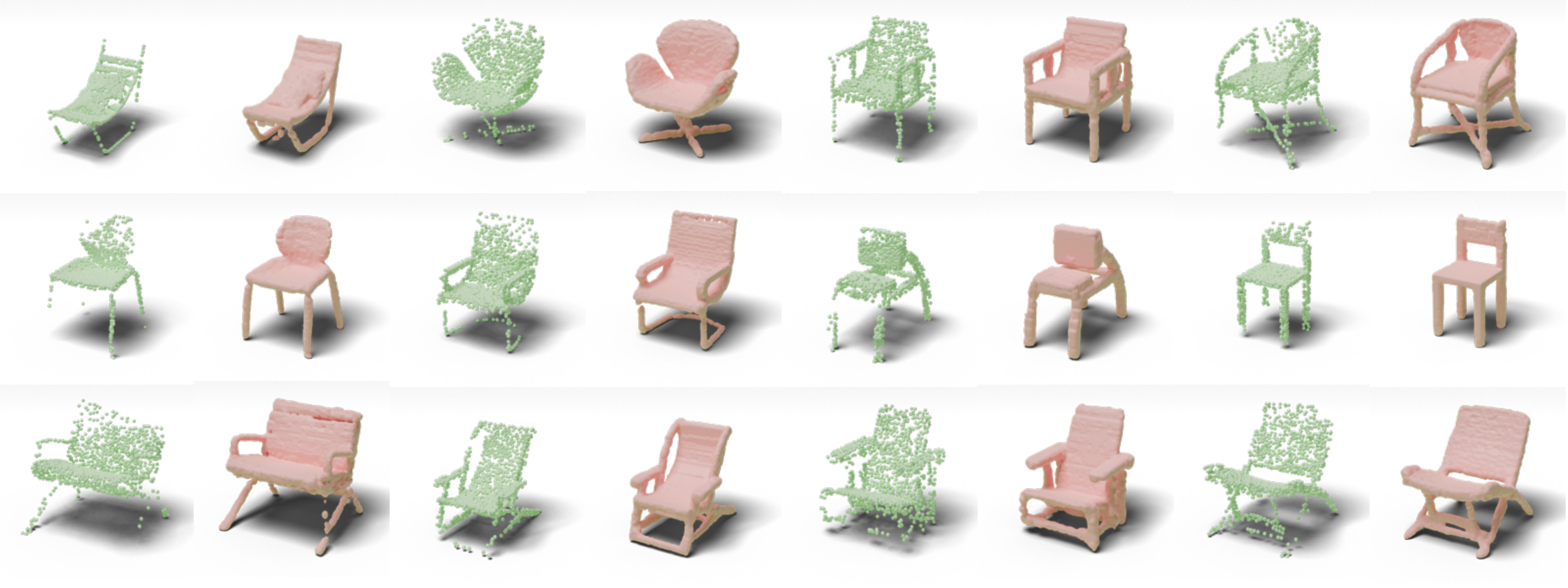}
\put(4,0) {Input}
\put(15.5,0) {Output}
\put(28.5,0) {Input}
\put(40.5,0) {Output}
\put(53.5,0) {Input}
\put(65,0) {Output}
\put(78,0) {Input}
\put(89,0) {Output}
\end{overpic}
\caption{More shape completion results on real scans for ScanNet\cite{dai2017scannet} objects. }
\label{Figure:supp_scan}
\end{figure*}

\begin{figure*}
\centering
\begin{overpic}[width=0.9\textwidth]{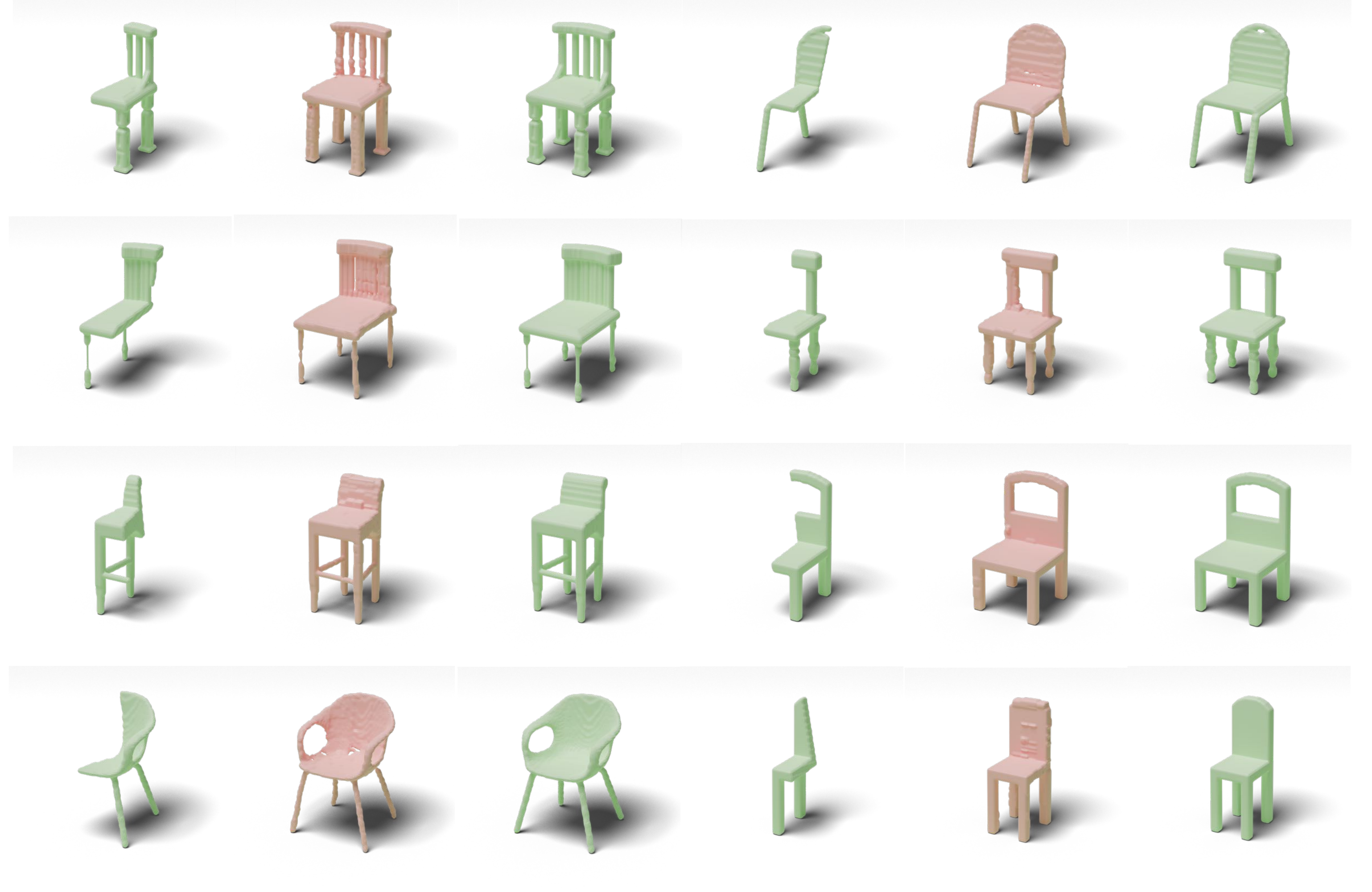}
\put(8,0) {Input}
\put(22,0) {Output}
\put(40,0) {GT}
\put(55,0) {Input}
\put(71,0) {Output}
\put(88.5,0) {GT}
\end{overpic}
\caption{More shape completion results on shapes with large missing areas.}
\label{Figure:supp_planecut}
\end{figure*}

\begin{figure*}
\centering
\begin{overpic}[width=0.95\textwidth]{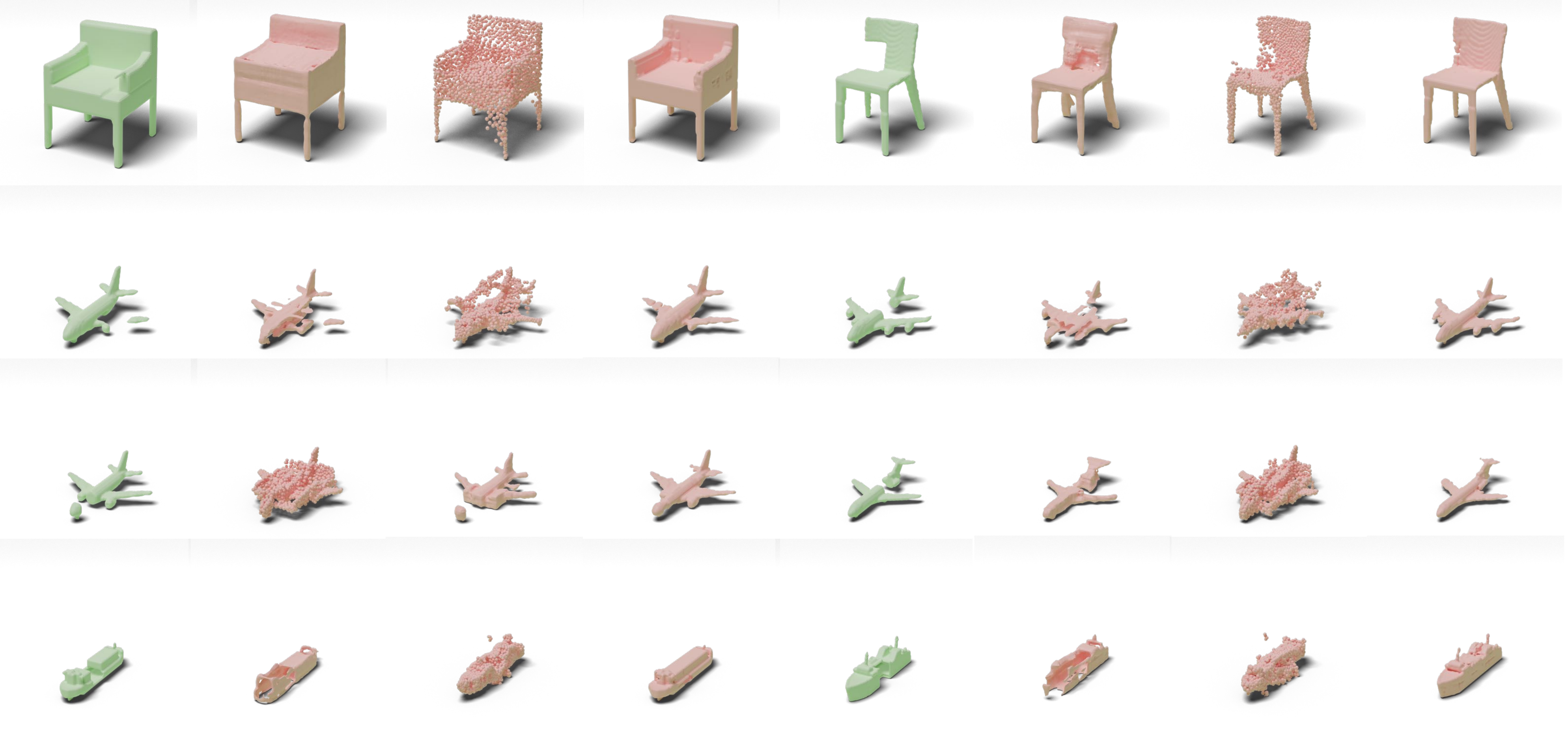}
\put(1,48) {Lamp $\rightarrow$ Chair}
\put(1,31.5) {Chair $\rightarrow$ Plane}
\put(1,20) {Cabinet $\rightarrow$ Plane}
\put(1,9) {Chair $\rightarrow$ Boat}
\put(2,-1) {Input}
\put(10,-1) {Conv-ONet\cite{Peng2020convonet}}
\put(28,-1) {Snow\cite{xiang21snow}}
\put(38,-1) {PatchRD}
\put(52,-1) {Input}
\put(60,-1) {Conv-ONet\cite{Peng2020convonet}}
\put(78,-1) {Snow\cite{xiang21snow}}
\put(89,-1) {PatchRD}
\end{overpic}
\caption{More testing results on novel categories. For each row, we note the training categories and testing categories on the left top corners. Lamp$\rightarrow$Chair means training on lamp and testing on chair shapes. }
\label{Figure:supp_cross}
\end{figure*}

We show more qualitative results of shape completion results on random-crop dataset (Figure \ref{Figure:supp_crop1} and Figure \ref{Figure:supp_crop2}), ScanNet\cite{dai2017scannet} objects (Figure \ref{Figure:supp_scan}), shapes with large missing areas (Figure \ref{Figure:supp_planecut}) and novel categories (Figure \ref{Figure:supp_cross}).

\section{More Results on PCN Benchmark}
\begin{figure}[h]
\centering
\begin{overpic}[width=\textwidth]{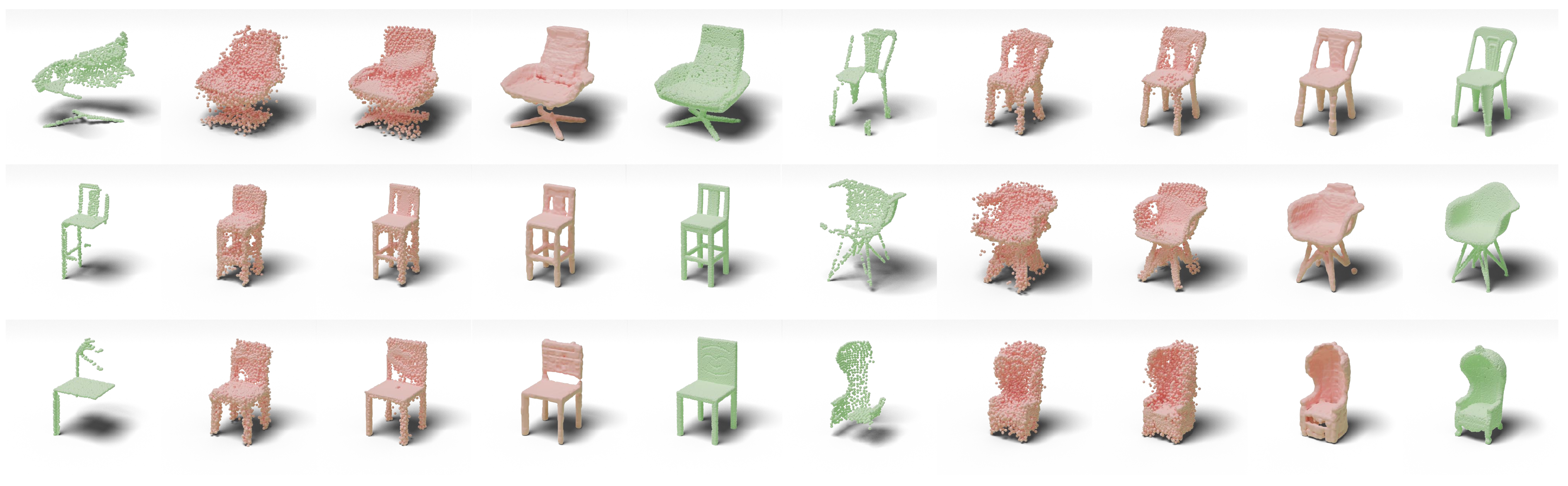}
\put(2,0) {  Input}
\put(11,0) { GRNet\cite{xie2020grnet}}
\put(22,0) { Snow\cite{xiang21snow}}
\put(31,0) { PatchRD}
\put(31.5,-2.1) { \textbf{(Ours)}}
\put(43,0) { GT}
\put(51,0) {  Input}
\put(60,0) { GRNet\cite{xie2020grnet}}
\put(71,0) { Snow\cite{xiang21snow}}
\put(80.5,0) { PatchRD}
\put(81.5,-2.1) { \textbf{(Ours)}}
\put(92,0) { GT}
\end{overpic}
\caption{More qualitative comparison on PCN Dataset\cite{yuan18pcn}. }
\label{Figure:supp_pcn}
\end{figure}

We show the quantitative results and more qualitative results on the PCN dataset \cite{yuan18pcn} in Table \ref{Table::supp::pcn} and Figure \ref{Figure:supp_pcn} respectively.
Our method is quantitatively a little worse than the best-performing SnowFlakeNet because our method might fail when there's no reference details in the input shape, and CD-L1 is more sensitive to structure than details. 
Importantly, visual results in Figure ~\ref{Figure:supp_pcn} indicate that our method produces more clean and plausible shapes, especially in the missing areas.


\section{More Training Details}

For the \textbf{coarse completion}, the input shape is the partial detailed shape and the ground truth is the coarse version ($4\times$ downsampled) of the detailed full shape. The loss function is the cross-entropy loss between the GT and the output. We use Kaiming Uniform method for weight initialization and the Adam optimizer to train 100 epochs for each shape category on a single Titan X. Training takes ${\sim}3$ hrs for the 3D CNN, ${\sim}12$ hrs for retrieval, and ${\sim}2$ hrs for deformation and blending. Inference for a shape with $128^3$ voxels takes ${\sim}20$s on a single 12GB Titan X.

\section{Failure Cases Analysis}
Our pipeline has 3 stages: (1)~coarse completion, (2)~patch retrieval, and (3)~patch deformation and blending. If one stage fails, the result might be different from the GT shape. However, our method can still produce plausible output, i.e. semantically correct and smoothly connected shapes. If stage 1 fails, the overall structure will be different from the GT shape. If stage 2 fails, the local details will be inaccurate. If stage 3 fails, the connection between patches will not be smooth, causing irregular or noisy shapes. 
Some examples of failure cases from each step are shown in Fig. \ref{Figure::supp::failure_steps}.

\section{Network Architectures}
\begin{figure}[h]
\centering
\begin{overpic}[width=0.4\textwidth]{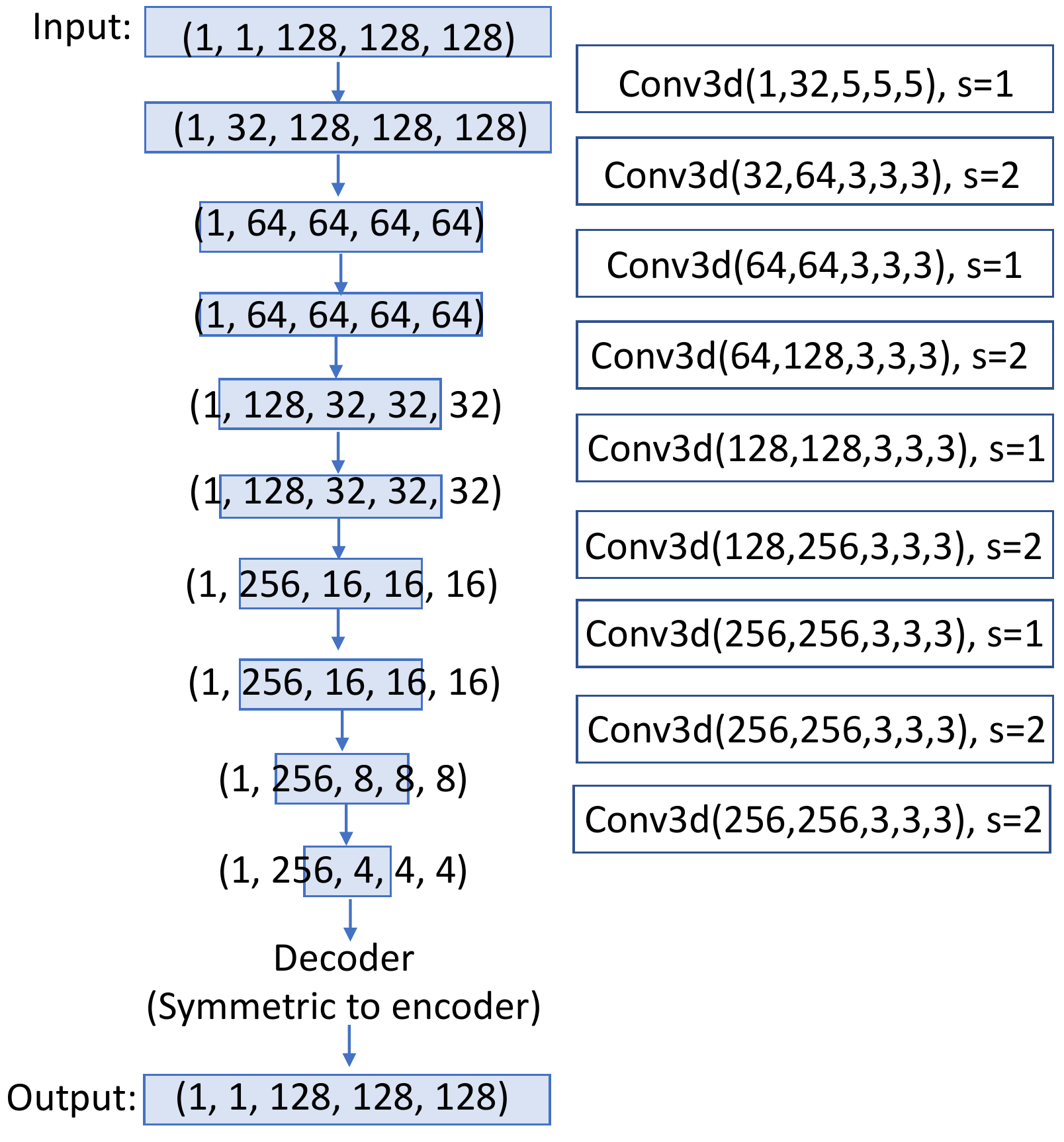}
\end{overpic}
\caption{Architecture of the coarse completion network. The input is a partial shape with size $(128,128,128)$ and the output is a coarse shape with the same size. We only show the encoder in detail here. The decoder is symmetric to the encoder. In the figure, blue boxes are tensors and white boxes are layers between two tensors.
The array after $Conv3d$ means (input channel, output channel, kernel size, kernel size,kernel size). $s$ means stride. }
\label{Figure:supp_completor}
\end{figure}

\begin{figure}[h]
\centering
\begin{overpic}[width=0.4\textwidth]{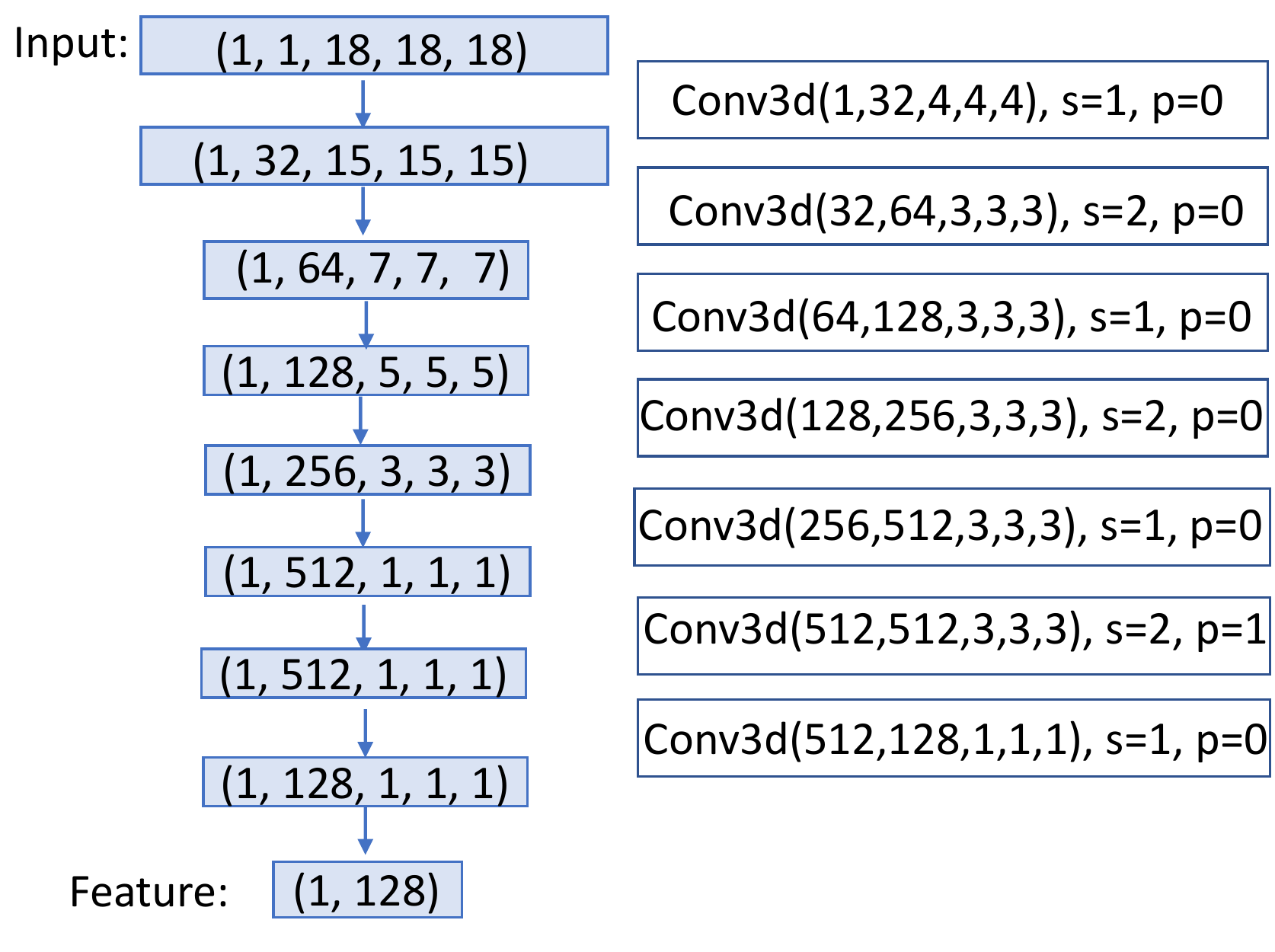}
\end{overpic}
\caption{Architecture of the feature encoder in the retrieval learning part. The input is a patch with size $(18, 18, 18)$. The output is a feature vector with size $128$.  In the figure, blue boxes are tensors and white boxes are layers between two tensors.
The array after $Conv3d$ means (input channel, output channel, kernel size, kernel size,kernel size). $s$ means stride. $p$ means padding. }
\label{Figure:supp_encoder}
\end{figure}

\begin{figure*}
\centering
\begin{overpic}[width=0.9\textwidth]{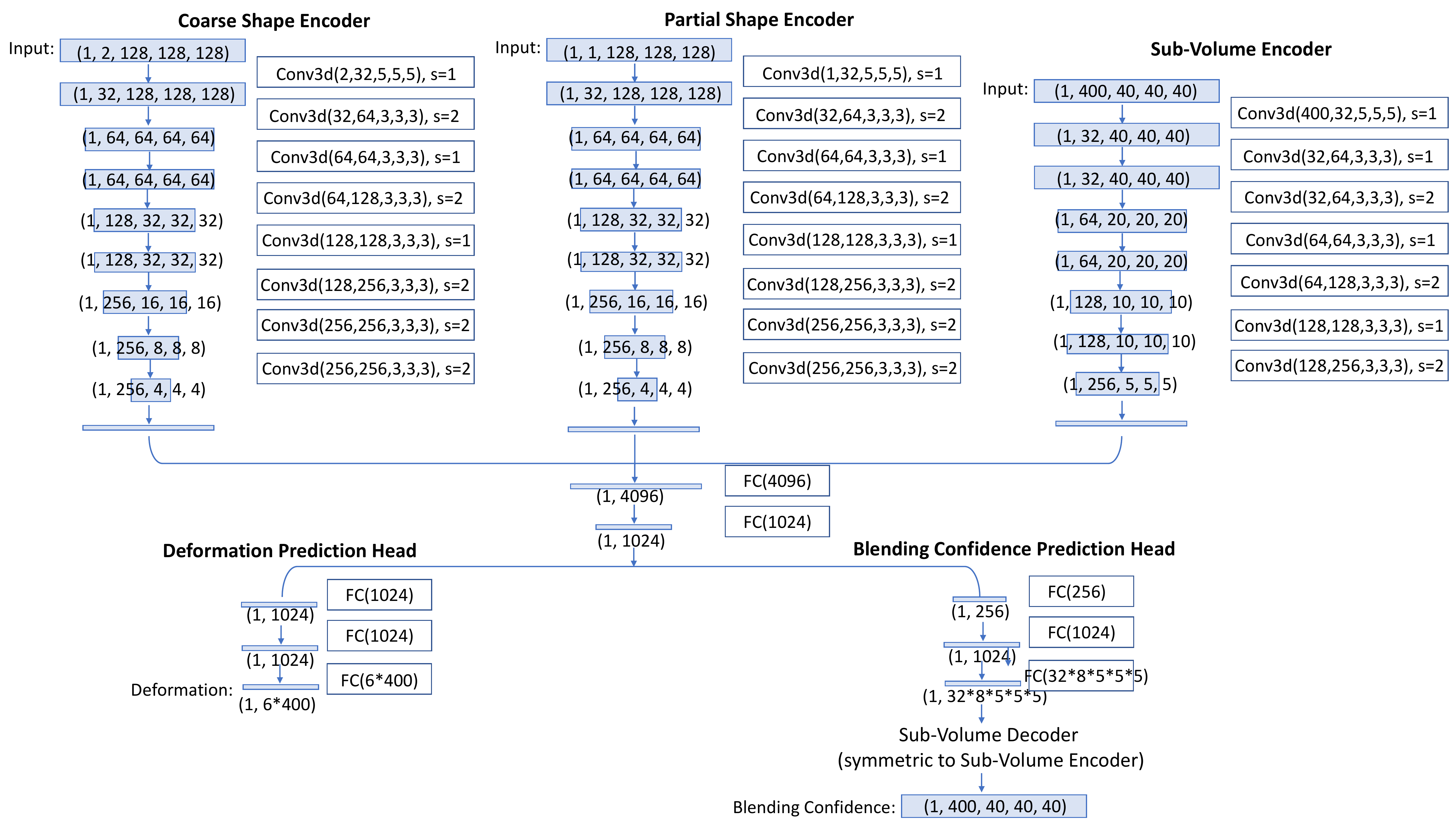}
\end{overpic}
\caption{Architecture of the patch deformation and blending weight prediction network. There are 3 branches to encode the coarse shape, partial shape and the sub-volume to one-dimensional feature vectors. Then two heads decode the concatenated feature vectors to deformation and blending weights. In the figure, blue boxes are tensors and white boxes are layers between two tensors.
The array after $Conv3d$ means (input channel, output channel, kernel size, kernel size,kernel size).  $s$ means stride.   }
\label{Figure:supp_deformer}
\end{figure*}

We show the detailed network architectures for coarse completion, retrieval metric learning, and deformation and blending weight prediction in Figure \ref{Figure:supp_completor}, Figure \ref{Figure:supp_encoder}, and Figure \ref{Figure:supp_deformer} respectively.

\begin{table}[h]
\centering
\begin{adjustbox}{width=0.6\columnwidth,center}
\begin{tabular}{ c | c c c c } 
 \hline
  & TopNet\cite{lyne19topnet} & GRNet\cite{xie2020grnet} & SnowFlakeNet\cite{xiang21snow} & PatchRD(Ours) \\
  \hline 
 CD-$L_1$ &13.43 &  9.37 & 7.78 & 8.79 \\
 \hline
\end{tabular}
\end{adjustbox}
\caption{Quantitative results on chair class of the PCN Dataset\cite{yuan18pcn}. All methods are trained on chair class only. We report the $L_1$ chamfer distance $\times 10^-3$. }
\label{Table::supp::pcn}
\end{table}

\begin{figure}[t!]
\centering
\includegraphics[width=0.8\linewidth]{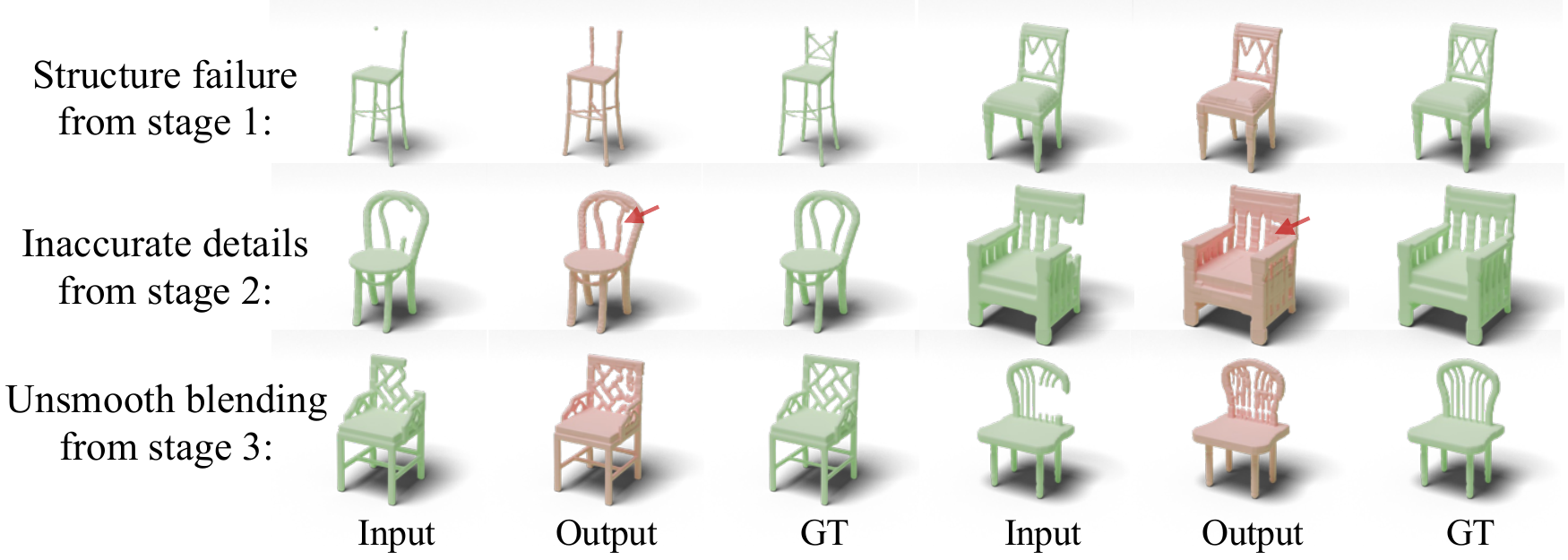}
\caption{Failure cases caused by different steps.}
\label{Figure:supp::failure_steps}
\end{figure}

